\documentclass[12pt,a4paper]{article}
\usepackage{amsfonts}
\usepackage{amsmath,amsfonts,amsthm,mathrsfs,grffile}
\usepackage{epstopdf}
\usepackage{graphicx,subfigure}
\usepackage{verbatim}
\usepackage{color}
\usepackage{picinpar}
\usepackage{booktabs}
\usepackage[doublespacing]{setspace}
\usepackage[authoryear]{natbib}

\newtheorem{thm}{Theorem}
\newtheorem{cor}{Corollary}
\newtheorem{lem}{Lemma}

\newtheorem{prop}{Proposition}

\theoremstyle{definition}   %

\theoremstyle{remark}

\numberwithin{equation}{section}

\newcommand{\RR}{\mathbb R}

\newcommand{\XX}{\mathbb X}

\newcommand{\be}{\mathbf{e}}

\newcommand{\ba}{\mathbf{a}}

\newcommand{\bbeta}{\boldsymbol{\beta}}

\newcommand{\bepsilon}{\boldsymbol{\epsilon}}

\newcommand{\bY}{\mathbf{Y}}
\newcommand{\bbf}{\mathbf{f}}
\newcommand{\KK}{\mathbb K}

\def\eop{{\hfill\vbox{\hrule height .3pt
      \hbox{\vrule width.3pt height 7pt
      \kern 7pt
      \vrule width .3pt}
      \hrule height .3pt}} \par\bigskip}

\title{ Debiased  distributed learning for sparse partial linear models in  high dimensions }

\author{Shaogao Lv$^{\dag,*}$, Heng Lian$^\ddag$
\\
{\small $^{\dag}$Department of Statistics and Mathematics,  Nanjing  Audit University, Nanjing, China;} \\
{\small $^{\ddag}$Department of Mathematics, City University of Hong Kong, Hong Kong;}\\
{\small $^*$Corresponding author: kenan716@mail.ustc.edu.cn }}
\date{}

\begin{document}
\maketitle

\setcounter{page}{1}
\begin{abstract}
Although various distributed machine learning schemes have been proposed recently for pure linear models and fully nonparametric models, little attention has been paid on distributed optimization for semi-paramemetric models with multiple-level structures (e.g. sparsity, linearity and nonlinearity). 
%Moreover, most of existing distributed work to handle big data don't consider the affect of the dimension of features, which is a crucial and challenging problem under the context of data parallelism. 
To address these issues, the current paper proposes a new communication-efficient distributed learning algorithm for partially sparse linear models with an increasing number of features. The proposed method is based on the classical divide and conquer strategy for handing big data and each sub-method defined on each subsample consists of  a debiased estimation of the double-regularized least squares approach. With the proposed method, we theoretically prove that our global parametric estimator can achieve optimal parametric rate in our semi-parametric model given an appropriate partition on the total data. Specially, the choice of data partition  relies on the underlying smoothness of the nonparametric component, but it is adaptive to the sparsity parameter. 
Even under the non-distributed setting, we develop a new and easily-read proof for optimal estimation of the parametric error in high dimensional partial linear model.
Finally, several simulated experiments are implemented to indicate comparable empirical performance  of our debiased technique under the distributed setting. 

%Our main contribution in the distributed framework lies in taking into interaction of different-type estimation with response to multiple-level structures in semi-parametric model and the affect of the number of dimension.
%for high dimensional partial linear models, where the sample with a total of $N$ data points is  randomly distributed  among $m$ machines and the parameters of interest are calculated by merging their $m$ individual estimators. This paper primarily focuses on the investigation of the high dimensional linear components in partial linear models, which is often of more interest.  We propose a new debiased averaging estimator of parametric coefficients on the basis of each individual estimator, and establish  new non-asymptotic oracle results in high dimensional and distributed settings, provided that $m\leq \sqrt{N/\log p}$ and other mild conditions are satisfied, where $p$ is the linear coefficient dimension. We also provide an experimental evaluation of the proposed method, indicating  the numerical effectiveness on simulated data. Even under the classical non-distributed setting,  we give the optimal rates of the parametric  estimator  with a looser tuning parameter limitation, which is required for our error analysis. 

\end{abstract}


\section{Introduction}
Under a big-data setting, the storage and analysis of data can no longer be performed on a single machine, and in this case dividing data into many sub-samples becomes a critical procedure for any numerical algorithm to be implemented.   Distributed statistical estimation and distributed optimization have received increasing attention in recent years, and  a flurry of research towards solving very large scale  problems have emerged recently, such as \citet{Mcdonald2009,Zhang2013, Zhang2015,Rosenblatt2016} and the references therein.  In general, distributed algorithm can be classified into two families:  data parallelism and task parallelism.  Data parallelism aims at  distributing the data across different parallel computing nodes or machines; and  task parallelism  distributes different tasks  across  parallel computing nodes. We are only concerned with data parallelism  in this paper. In particular, we primarily consider the distributed estimation for  partially linear models via using the standard divide and conquer strategy.  {\it Divide-and-conquer} technology is a simple and communication-efficient way for handling big data, which is commonly used in the literature of statistical learning. To be precise,  the whole data is randomly allocated among $m$ machines, a local estimator is computed independently on each machine, and then the central node averages the local solutions into a global estimate.

Partially linear models (PLM)  \citep*{Hardle2007,Heckman1986}, as the leading example of semiparametric models, are a class of important tools for modeling complex data, which retain model interpretation and flexibility simultaneously. 
Given the observations $(Y_i,X_i,T_i), i=1,...,N$, where $Y_i$ is the response, $X_i=(x_{i1},...,x_{ip})\in\RR^p$ and $T_i=(t_{i,p+1},....,t_{i,p+q})\in\RR^q$ are
vectors of covariates, the partially linear models assume that
\begin{equation}\label{gmodel}
 Y_i=X_i'\bbeta^*+f^*(T_i)+\epsilon_i,
\end{equation}
where  $\bbeta^*\in \RR^p$ is a vector of unknown parameters for linear terms, $f^*$ is an unknown function defined on a compact subset $\mathcal{T}$ of $\mathbb{R}^q$ ($q$ is fixed), and $\epsilon_i$'s are independently standard  normal variables. In the sparse setting, one often assumes that the cardinality of nonzero components of $\bbeta^*$ is far less than $p$, that is,
$s^*:=|S:=\{j\in[p],\,\beta^*_j\neq 0\}|\ll p$.

There is  a substantial body of work focusing on the sparse setting for PLM, see, for example, \citet{Green1985,Wahba1990,Hardle2007,Zhang2011,Lian2012,Wang2014}, among others. \cite{Chen1988} and \cite{Robinson1988} showed that the parametric part can be estimated with parametric rates under appropriate conditions. \cite{Mammen1997} proved the linear part is asymptotically normal under a more general setting. These results are asymptotic and valid in the fixed dimensions, where the number of variables $p$ in the linear part is far less than the number of observations.

Although this paper is mainly concerned with data parallelism, which is practically useful in the $N>p$ setting, the size of each sub-sample $(n:=N/m)$ that is allocated to each node may be less than $p$ with a large number of nodes ($m$). So the high dimensional issue has been endowed with additional implications under the data parallelism setting. Compared to linear models or nonparametric additive models, the high dimensional case for studying PLM with $p>n$ is more challenging, mainly because of the correlation and interaction effect between covariates in the linear part  and covariates in the nonparametric part. 
Under the high dimensional framework, a commonly-used approach is to construct penalized least squares estimation with a double penalty terms, using a smoothing functional norm to control complexity of the nonparametric part and a shrinkage penalty on parametric components to achieve model parsimony. To build each individual estimator before merging data, we consider a double-regularized approach with the Lasso penalty and a reproducing kernel Hilbert space (RKHS) norm,  given by  
\begin{equation}\label{submethod}
\min_{\bbeta\in\RR^p,f\in \mathcal{H}_K}\{\mathcal{L}^{(l)}(\bbeta, f)\},\,\,\,
\mathcal{L}^{(l)}(\bbeta, f)=\frac{1}{2n}\sum_{i\in S_l}\big(Y_i-X_i'\boldsymbol{\beta}-f(T_i)\big)^2+\lambda_1\|\bbeta\|_1+
\lambda_2/2 \|f\|_{K}^2,
\end{equation}
where $S_l$ is the $l$-th subsample with data size $n$, and $(\lambda_1,\lambda_2)
$ are two tuning parameters.
Here we consider a function from a RKHS denoted by $\mathcal{H}_{K}$, endowed with the norm $\|\cdot\|_{K}$. The kernel function $K$ defined on $\mathcal{T}\times \mathcal{T}$ and $\mathcal{H}_{K}$ is determined by each other. 

With diverging dimensions in the linear part, there is a rich literature on penalized estimation for PLM in the last decade. \citet*{Xie2009} proposed the SCAD-penalized estimators of the linear coefficients, and achieved estimation consistency and variable selection consistency  for the linear and nonparametric components. Similar to \eqref{submethod}, \cite{Ni2009} formulated a double-penalized least squares approach, using the smoothing spline to estimate the nonparametric part and the SCAD to conduct variable selection. It is shown that the proposed procedure can be as efficient as the oracle estimator. Recently, \cite{Wang2014}  proposed a new doubly penalized procedure for variable selection with respect to both linear and additive components, where  the numbers of linear covariates and nonlinear components both diverges as the sample size increases. All these aforementioned papers focus on the case where $p$ is relatively small compared to $n$ $(e.g.\,\,\, p=o(\sqrt{n}))$.

Allowing for $p\geq n$ and even $p=o(e^n)$, recent years has witnessed several related research in terms of non-asymptotic analysis for the sparse PLM. As shown in the distributed learning literature \citep*{Lee2015,Zhang2015}, the optimal estimation of each local estimate is very critical to derive the optimal non-asymptotic results of the averaging estimate. Under the non-distributed setting, \citet*{Muller2015} theoretically analyzed the penalized estimation \eqref{submethod}, and proved that the parametric part in PLM achieves the optimal rates of the linear models, as if the nonparametric component were known in advance. More recently, \cite{Zhu2017} considered a two-step approach for estimation and variable selection in PLM. The first step  uses nonparametric regression to obtain the partial residuals, and the second step is an $\ell_1$-penalized least squares estimator (the Lasso) to fit the partial residuals from the first step. Like \citet*{Muller2015}, they derived  optimal non-asymptotic oracle results for  the linear estimator. 
% However, we find that the above mentioned results cannot be used directly in our case. In fact, the optimal oracle results established in \cite{Muller2015} can be achieved  only when  the decay of the tuning parameter $\lambda_2$  is not  faster than $n^{-\frac{2\alpha}{2\alpha+1}}$, where $\alpha$ is a nonparametric complexity.
% This constraints on  $\lambda_2$  makes it impossible to obtain the optimal rates of the averaging estimate.  
%Recently, \cite{Zhu2017} considered a two-step approach for PLM which differs from our one-step penalized estimation. 
%To this end, under more relaxed constraints on the tuning parameters, we prove theoretical results for both the prediction and the estimation error. 

In this paper, we aim at proposing an efficient distributed estimation for high dimensional PLM. Although distributed estimation on linear models \citep*{Zhang2013,Lee2015,Battey2015} and on fully nonparametric models \citep*{Zhang2015,Lin2016} have been well-understood, the investigation on distributed estimation for PLM is more challenging and there are very few works on this \citep*{Zhao2016,Lian2019}. 
First, it is known that the Lasso penalty and the functional norm in \eqref{submethod} lead to a heavily biased estimation, and naive averaging only reduces the variance of the local estimators, but has no effect on the bias \citep*{Mcdonald2009, Wu2017}. Moreover, in the diverging dimensional setting (i.e. $p, n\rightarrow \infty$), \citet*{Rosenblatt2016} showed that the averaged empirical risk minimization (ERM) is suboptimal versus the centralized ERM. So debiasing is essential to improving accuracy of the averaging estimate.  In addition, the significant influences of high dimensions and correlated covariates from the parametric and  nonparametric components will result in additional technical difficulties.

Our main contribution to this line of research consists of  two aspects. Our first contribution is to analyze the double-regularized least squares method \eqref{submethod} for estimating sparse PLM and provide upper bounds on the parametric part and the nonparametric part respectively. These derived results based on any given subsample  serve our proposed averaging estimation merging the total data. Our proof for optimal rates of the parameter estimator in PLM is partially inspired by the idea from \citep*{Muller2015}, but there exists some technical distinctions from the main proof in \citep*{Muller2015} under the non-distributed setting.
In particular, the learning rate of the parametric estimator is based on the zero-order optimization, while
the corresponding proof in Theorem 2 of \citep*{Muller2015} uses the first-order optimization. 
The latter may  be only applicable to those strongly convex learning problems, and thereby excludes the Lipschizt-loss based learning, such as the classical SVM and the quantile regression. It is known that the zero-order optimization is sufficient for establishing optimal 
estimation consistency, while the first order information of model estimation is not essential to estimation consistency.
See details for high dimensional linear models  \citep*{Bickel2009} and  full non-parametric regression model \citep*{Raskutti2012}. To the best of our knowledge, our proof is the first one that provide optimal estimation error
for the high dimensional PLM only using the zero-order optimization.
%rates of various estimators forfor the linear models and. In other words, the first order information of learning approaches is not needed for estimation consistency. Our new proof for proving estimation error shows that these conclusions also holds for high dimensional PLM 
 %Recall that optimal rates for  estimating high dimensional linear models and full non-parametric models have been established  when the linear coefficients are restricted to a bounded domain, the optimal oracle rates of the parametric estimator can be achieved using a broad range of values of $\lambda_2$, which thereby yields  the corresponding optimal bounds for the debiased averaging estimator. 
 
  Note that from the proof on our debiased averaging estimation, it is seen that  the nonparametric  component in PLM significantly affects estimation error of the averaging parametric estimator. Hence, error bound of our distributed parametric estimator also depends on the functional complexity of nonparametric components. This observation
   differs from  these existing results obtained under the non-distributed setting \citep*{Hardle2007,Muller2015}. Theorem 1 indicates that, under the ultra-high dimensional setting (where the error of parametric estimation dominates the nonparametric one), the parametric estimation with the optimal rate can be guaranteed with  
   an appropriate splitting number $m$, which does not depend on the complex parameter of the nonparametric component. Otherwise, the optimal parametric rate is also guaranteed with a smoothness-dependent  $m$.
  % derived upper bound for parameter estimation for the high dimensional PLM is optimal in the  minimax sense.
  % Only by choosing a sufficiently small $\lambda_2$, the averaging estimate can achieve the same rates as the centralized M-estimator based on the entire data. This is why the mentioned previous results \citep*{Muller2015} are not applicable in our analysis.
%Even without any boundedness restriction of $\bbeta$, the optimal rates of the nonparametric estimation are still obtained, while in this case we only obtain  suboptimal rates for the parametric part.

Our second contribution is to propose a novel debiased distributed estimation for the sparse PLM under the big-data setting, in that simply averaging cannot reduce the estimation bias in contrast to the local estimators. To our knowledge,  this is the first work that considers distributed problems on the high dimensional PLM. In fact, our study in this paper is related to the previous work \citep*{Zhao2016}, where they considered the naive averaging strategy for the PLM with non-sparse coefficients and fixed dimensions. So their work differs from the current paper in terms of problem setup and methodological strategy. Although the debiasing technology has been employed for the sparse linear regression \citep*{Lee2015},  analyzing  the  debiased distributed estimation for PLM is more challenging, mainly because thenonparametric component affects error level of the averaging estimation. To handle this problem, we apply some abstract operator theory to provide upper bounds of this approximation error in  RKHS. By contrast, some existing related results \citep*{Wahba1990,Ni2009} depend on some strong assumptions on data sampling, for example, $T_i$'s are deterministically drawn from $[0,1]$ such that $\int_0^{T_i}u(t)d(t)=i/n$, where $u(t)$ is a continuous and positive function. From our simulated results,  we see the estimation error of the averaging debiased parametric estimator is comparable to that of the centralized M-estimator, while that of the naive averaged parametric estimator is much worse.

The rest of the paper is organized as follows. In Section 2, we provide some background on kernel spaces and propose a debiased averaging parametric estimator based on each local estimate \eqref{submethod}. Section 3 is devoted to the statement of our main results and discussion of their consequences. In Section 4, for  local parametric estimation, we present general upper bounds on the estimation error. Section 5 contains the technical details, and some useful lemmas are deferred to the Appendix. Some simulation experiments are reported in Section 6, and we conclude in Section 7.

 {\bf Notations}. In the following, for a vector $Z=(z_1,...,z_p)'$, we use $\|Z\|_1$, $\|Z\|_2$ to represent $\ell_1$ and $\ell_2$-norm in the Euclidean space respectively, and also
$\|Z\|_\infty=\max_{j\in[p]}|z_j|$. For a matrix $A$, $\|A\|_2$ denotes the spectral norm. For a function $f$ defined on $\mathcal{Z}$ and a given data $(Z_i)_{i=1}^n$ drawn from the underlying distribution $\rho$ defined on $\mathcal{Z}$, let $\|f\|_2:=\sqrt{\mathbb{E}[f^2(Z)]}$ be the $L_2(\rho)$-norm for any square-integrable function $f$. We use $\|f \|_n$ to denote the empirical $L_2(\rho)$-norm, i.e. $\|f\|_n^2=\frac{1}{n}\sum_{i=1}^nf(Z_i)^2$. For sequences $f(n),\,g(n)$, $f(n)=\Omega(g(n))$ means that there is some constant $c$, such that $f(n)\geq cg(n)$, and  $f(n)=O(g(n))$ means that $f(n)\leq c'g(n)$ for an absolute constant $c'$ with probability approaching one. The symbols  $C,\,c$ with various subscripts are used to denote different constants. For $q\in \mathbb{N}^+$,  we write $[q]:=\{1,...,q\}$.

\section{Background and The Proposed Estimator}
We begin with some background on RKHSs, and then formulate a profiled Lasso-type approach equivalent to the double-regularized one \eqref{submethod}. Based on a gradient-induced debiasing and an estimate to approximate the inverse weighted covariate matrix, we propose the debiased averaging parametric estimator for PLM.  

\subsection{Reproducing kernel Hilbert space}
Given a  subset $\mathcal{T}\in \RR^q$ and a probability measure $\rho_T$, we define a symmetric non-negative kernel function $K:\,\mathcal{T}\times \mathcal{T}\rightarrow \RR$, associated with the RKHS of functions from $\mathcal{T}$ to $\RR$. The Hilbert space $\mathcal{H}_K$ and its dot product $\langle \cdot, \cdot \rangle_K$ are such that (i) for any $t\in \mathcal{T}$, $K_t(\cdot):=K(\cdot, t)\in \mathcal{H}_K$; (ii) the reproducing property holds, e.g. $f(t)=\langle f, K_t\rangle_K$ for all $f\in \mathcal{H}_K$. It is known that the kernel function $K$ is determined by $\mathcal{H}_{K}$ \citep*{Aronszajn1950}. Without loss of generality, we assume that $\kappa:=\sup_{t\in \mathcal{T}}|K(t,t)|\leq 1$, and such a condition includes the Gaussian kernel and the Laplace kernel as special cases.

The reproducing property of RKHS plays an important role in theoretical analysis and numerical optimization for any kernel-based method. Specially, this property implies that $\|f\|_\infty\leq \kappa \|f\|_K\leq \|f\|_K$ for all $f\in\mathcal{H}_K$. Moreover, by Mercer's theorem, a kernel $K$ defined on a compact subset $\mathcal{T}$ admits the following eigen-decomposition:
$$
K(t,t')=\sum_{\ell=1}^\infty \mu_\ell \phi_\ell(t) \phi_\ell(t'),\,\,t,t'\in \mathcal{T},
$$
where $\mu_1\geq \mu_2\geq \cdots>0$ are the eigenvalues and $\{\phi_\ell\}_{\ell=1}^\infty$ is an orthonormal basis in $L_2(\rho_T)$. The decay rate of $\mu_\ell$ fully characterizes the complexity of the RKHS induced by the kernel $K$, and has close relationships with various entropy numbers, see \citet*{Steinwart2008}
for details. Based on this, we define the quantity:
$$
\mathcal{Q}_n(r)=\frac{1}{\sqrt{n}}\Big[\sum_{\ell=1}^{\infty}\min\{r^2,\mu_\ell\}\Big]^{1/2},\quad \forall \,r>0.
$$
Let $\nu_n$ be the smallest positive solution to the inequality:
$
40\nu_n^2\geq\mathcal{Q}_n(\nu_n),
$
where $40$ is only a technical constant. Then, due to the high dimensional effect on the nonparametric estimation for PLM, we  introduce the following quantity related to the convergence rates of semi-parametric estimate:
$$
\gamma_n:=\max\Big\{\nu_n,\sqrt{\frac{\log p}{n}} \Big\}.
$$

\subsection{The Debiased  Estimator}

For the $l$-th machine, define $\XX^{(l)}=(X_1^{(l)},...,X_n^{(l)})'$, $\bepsilon^{(l)}=(\epsilon_1^{(l)},...,\epsilon_n^{(l)})'$, $\bY^{(l)}=(Y_1^{(l)},...,Y_n^{(l)})'$ and $\bbf^{(l)}=(f(T_1^{(l)}),...,f(T_n^{(l)}))'$.
$\KK^{(l)}$ is a semi-definite $n\times n$ matrix whose entries are $(K(T_i^{(l)},T_j^{(l)}))_{i,j=1}^n$.
The partially linear model \eqref{gmodel} can then be written as $\bY^{(l)}=\XX^{(l)}\bbeta^*+(\bbf^*)^{(l)}+\bepsilon^{(l)}$. By the reproducing property of RKHS \citep*{Aronszajn1950}, the nonparametric minimizer of programme \eqref{submethod} has the form $f=\sum_{i\in S_l}a_i K(X_i,\cdot)$ and particularly $\bbf^{(l)}=\KK^{(l)}\ba$. Hence we can write $\mathcal{L}^{(l)}(\bbeta, f)$ as
\begin{equation}
\mathcal{L}^{(l)}(\bbeta, \ba):=\frac{1}{2n}\|\bY^{(l)}-\XX^{(l)}\bbeta-\KK^{(l)}\ba\|_2^2+\lambda_1\|\bbeta\|_1+\frac{\lambda_2}{2}\ba^T\KK^{(l)}\ba.
\end{equation}
Given $\lambda_1,\lambda_2$ and $\bbeta$, the first order optimality condition for convex optimization \citep*{Boyd2004} yields the solution
\begin{equation}\label{nonpaexpre}
\hat{\ba}^{(l)}=(n\lambda_2 \mathbb{I}+\KK^{(l)})^{-1}(\bY^{(l)}-\XX^{(l)}\bbeta),
\end{equation}
and $\mathbb{A}^{(l)}(\lambda_2)=\KK^{(l)}(\lambda_2 n\mathbb{I}+\KK^{(l)})^{-1}$ is equivalent to the linear smoother matrix in \citep{Heckman1986}. Indeed, $\mathbb{A}^{(l)}(\lambda_2)$ can be replaced by arbitrary smoother for specific purposes. Plugging $\hat{\bbf}^{(l)}=\KK^{(l)}\hat{\ba}^{(l)}$ into \eqref{submethod},
we can obtain a penalized problem only involving  $\bbeta$:
 \begin{equation}\label{lasso}
\mathcal{Q}^{(l)}(\bbeta):=\frac{1}{2n}(\bY^{(l)}-\XX^{(l)}\bbeta)'\big(\mathbb{I}-\mathbb{A}^{(l)}(\lambda_2)\big)(\bY^{(l)}-\XX^{(l)}\bbeta)
+\lambda_1\|\bbeta\|_1,
\end{equation}
and the quadratic term in $\mathcal{Q}^{(l)}(\bbeta)$ is called the profiled least squares in the literature.
Note that $\mathbb{I}-\mathbb{A}^{(l)}(\lambda_2)=\big( \mathbb{I}+\mathbb{K}_t^{(l)}/(\lambda_2n)\big)^{-1}$ is a nonnegative definite smoothing matrix.

Since the gradient vector of the empirical risk  $\frac{1}{2n}(\bY^{(l)}-\XX^{(l)}\bbeta)'\big(\mathbb{I}-\mathbb{A}^{(l)}(\lambda_2)\big)(\bY^{(l)}-\XX^{(l)}\bbeta)$ at $\hat{\bbeta}$ is
$$\frac{1}{n}(\XX^{(l)})'\big(\mathbb{I}-\mathbb{A}^{(l)}(\lambda_2)\big)(\bY^{(l)}-\XX^{(l)}\hat{\bbeta}),$$
which is just a sub-gradient of $\lambda_1\|\cdot\|_1$ at $\hat{\bbeta}$ by the classical KKT conditions. 
By adding a term proportional to the sub-gradient of the empirical risk for debiasing,  any debiased Lasso estimator compensates for the bias incurred by regularization. To be precise, motivated by the idea in
 in \citet*{Javanmard2014},
the debiased estimator from the $l$-th subsample with respect to the Lasso estimator $\hat{\bbeta}^{(l)}$ is given by
$$
\check{\bbeta}^{(l)}:=\hat{\bbeta}^{(l)}+\frac{1}{n}\hat{\Theta}^{(l)}(\XX^{(l)})'\big(\mathbb{I}-\mathbb{A}_u^{(l)}(\lambda_2)\big)(\bY^{(l)}
-\XX^{(l)}\hat{\bbeta}^{(l)}),
$$
where $\hat{\Theta}^{(l)}$ is an approximate inverse to the weight empirical covariance matrix $\tilde{\Sigma}_l:=\frac{1}{n}(\tilde{\XX}^{(l)})'\tilde{\XX}^{(l)}$ on the design matrix $\tilde{\XX}^{(l)}:=\big(\mathbb{I}-\mathbb{A}_u^{(l)}(\lambda_2)\big)^{1/2}\XX$. Here $\mathbb{I}-\mathbb{A}_u^{(l)}(\lambda_2):=\big( \mathbb{I}+\mathbb{K}_t^{(l)}/(\lambda_2)\big)^{-1}$ is viewed as an unnormalized version of $\mathbb{I}-\mathbb{A}^{(l)}(\lambda_2)$ for technical conveniences.
Note that we drop the dependence on $\lambda_2$ of $\tilde{\XX}^{(l)}$ for simplicity. 
% $\tilde{\XX}^{(l)}$ can be regarded as an approximate  matrix  of
%the original design matric $\XX^{(l)}-\mathbb{E}[\XX^{(l)}|T]$, which has been frequently used in semi-parametric models \citep*{Xie2009}.
 By the debiasing technique,
 \citet*{Javanmard2014} proved that the bias of the debaised estimator is of the smaller order  than its variance, thus statistical inference such as asymptotic normality can be tractable.
s

Note also that the choice of $\hat{\Theta}^{(l)}$ is crucial to the performance of
the debiased estimator, and some feasible algorithms for forming $\hat{\Theta}^{(l)}$ has been proposed  recently by \citet*{Cai2011}, \citet*{Javanmard2014} and \cite{Geer2014}.
Thus, the averaged parametric estimator $\bar{\bbeta}$ by combining  the debiased estimators from all the subsamples
is given by
\begin{equation}\label{avergingestim}
\bar{\bbeta}=\frac{1}{m}\sum_{l=1}^m \check{\bbeta}^{(l)}.
\end{equation}

In this paper, we employ an estimator for forming   $\hat{\Theta}^{(l)}$ proposed by \cite{Geer2014}: nodewise
regression on the predictors. More precisely, for some $j\in [p]$, the $l$-th machine solves
\begin{equation}\label{covsolver}
\hat{\theta}_j:=\arg \min_{\theta \in \mathbb{R}^{p-1}}\frac{1}{2n}\|\tilde{X}_{l,j}-\tilde{\mathbb{X}}_{l,-j}\theta\|_2^2+
\lambda^{(j)}\|\theta\|_1,
\end{equation}
where $\tilde{\mathbb{X}}_{l,-j}\in \mathbb{R}^{n\times(p-1)}$ is $\tilde{\mathbb{X}}^{(l)}$ less its $j$-th column $\tilde{X}_{l,j}$.
Then we can define a non-normalized  covariance matrix by
\begin{equation*}
\hat{C}_l:=\left[
%\begin{equation*}
\begin{array}{cccc}
 1 & -\hat{\theta}_{1,2} & \cdots & -\hat{\theta}_{1,2} \\
-\hat{\theta}_{2,1} & 1 & \cdots & -\hat{\theta}_{2,p} \\
 \vdots & \vdots & \ddots & \vdots \\
 -\hat{\theta}_{p,1}& -\hat{\theta}_{p,2}& \cdots &1
\end{array}
\right],
\end{equation*}
where $\hat{\theta}_{j,k}$ is  the $k$-th element of $\hat{\theta}_j$, indexed by $k\in\{1,...,k-1,k+1,...,p\}$.
To scale the rows of $\hat{C}_l$, we define a diagonal matrix $\hat{T}_l:=\hbox{diag}(\hat{\tau_1},...,\hat{\tau_p})$ by
$$
\hat{\tau_j}=\big(\frac{1}{n}\|\tilde{X}_{l,j}-\tilde{\mathbb{X}}_{l,-j}\hat{\theta}_j\|_2^2+
\lambda^{(j)}\|\hat{\theta}_j\|_1\big)^{\frac{1}{2}},
$$
and based on this, we form  $\hat{\Theta}^{(l)}=\hat{T}_l^{-2}\hat{C}_l$.

%Under some regualizty conditions, Theorem 2.4 in \cite{Geer2014} has shown that
%$$
%\big\|\hat{\Theta}_j-\Sigma_j^{-1}\big\|_1 \preceq \Big(\frac{s_j^2\log p}{n}\Big)^{\frac{1}{2}},\,\,\forall\,
%j\in [p],
%$$
%where $\lambda_1_j\simeq \big(\frac{\log p}{n}\big)^{\frac{1}{2}}$ is chosen and
%$\Sigma$ is the weight population covirance matrix associated with  $\tilde{\Sigma}$, and
%$s_j$ is the sparsity of $\Sigma_j^{-1}$.

In order to show that $\hat{\Theta}^{(l)}$ is an approximate inverse of $\tilde{\Sigma}_l$, the first order optimality conditions of \eqref{covsolver}
 is applied to yield
\begin{align*}
\hat{\tau_j}^2&=\frac{1}{n}\|\tilde{X}_{l,j}-\tilde{\mathbb{X}}_{l,-j}\hat{\theta}_j\|_2^2+
\lambda^{(j)}\|\hat{\theta}_j\|_1\\
&= \frac{1}{n}\|\tilde{X}_{l,j}-\tilde{\mathbb{X}}_{l,-j}\hat{\theta}_j\|_2^2+\frac{1}{n}\big(\tilde{X}_{l,j}-\tilde{\mathbb{X}}_{l,-j}\hat{\theta}_j\big)^T
\tilde{\mathbb{X}}_{l,-j}\hat{\theta}_j\\
&=\frac{1}{n}\big(\tilde{X}_{l,j}-\tilde{\mathbb{X}}_{l,-j}\hat{\theta}_j\big)'
\tilde{X}_{l,j},\,\,\,j\in[p].
\end{align*}
Let $\hat{\Theta}^{(l)}_{j,.}$ be the $j$-th row of $\hat{\Theta}^{(l)}$,
according to the definition of $\hat{\Theta}^{(l)}$, it follows from the above equality that
\begin{align*}
\frac{1}{n}\hat{\Theta}^{(l)}_{j,.}(\tilde{\XX}^{(l)})'\tilde{X}_{l,j}
=\frac{1}{n\hat{\tau_j}^2}\big(\tilde{X}_{l,j}-\tilde{\mathbb{X}}_{l,-j}\hat{\theta}_j\big)'
\tilde{X}_{l,j}=1,\,\,\,j\in[p],
\end{align*}
and the optimality condition of \eqref{covsolver} is applied again to get
\begin{align*}
\frac{1}{n}\big\|\hat{\Theta}^{(l)}_{j,.}(\tilde{\XX}^{(l)})'\tilde{\XX}^{(l)}_{-j}\big\|_\infty
=\frac{1}{\hat{\tau_j}^2}\big\|\frac{1}{n}\big(\tilde{X}_{l,j}-\tilde{\mathbb{X}}_{l,-j}\hat{\theta}_j\big)'\tilde{\XX}^{(l)}_{-j}\big\|_\infty
\leq \frac{\lambda^{(j)}}{\hat{\tau_j}^2},\,\,\,j\in[p].
\end{align*}
Finally, we have
\begin{equation}\label{inverseappro}
\max_{j\in[p]}\big\|\hat{\Theta}^{(l)}_{j,.}\tilde{\Sigma}_l-\mathbf{e}_j\big\|_\infty\leq \max_{j\in[p]}\big\{\lambda^{(j)}/\hat{\tau_j}^2\big\}.
\end{equation}

We remark that forming $\hat{\Theta}^{(l)}$ is $p$-times as expensive as solving the local Lasso problem, which is the most expensive step of evaluating the averaging estimator. To this end, we could consider an estimator only using a common $\hat{\Theta}$ for all the local estimators in the following way. To reduce computational cost, we assign the task of computing  $p/m$ rows of $\hat{\Theta}$ to each local machine. Then each machine sends $\frac{p}{m}$ rows of $\hat{\Theta}$ computed by \eqref{covsolver} to the central server, as well as sending $\hat{\bbeta}^{(l)}$ and $(\XX^{(l)})'\big(\mathbb{I}-\mathbb{A}_u^{(l)}(\lambda_2)\big)(\bY^{(l)}-\XX^{(l)}\hat{\bbeta}^{(l)})$. Here we use different  $\hat{\Theta}^{(l)}$ for different machine merely for convenience of presentation and implementation.

We also remark that, allowing for moderately high dimensional case,  similar arguments as 
Lemma 1 in \cite{Speckman1988} with \eqref{inverseappro} tell us that $\hat{\tau_j}$'s converges to the corresponding eigenvalues of $\Xi$. Here $\be=X-\mathbb{E}[X|T]$ and we write $\Xi:=\mathbb{E}[Var(\be|T)]$. 
Moreover,
the positive definiteness of $\Xi$ is a standard assumption for obtaining semi-parametric efficient estimation, see \citep*{Speckman1988,Xie2009,Zhao2016}. In ultra-high dimensional setting, the underlying sparse structure of $\Xi$ is required to ensure consistency. We impose a related assumption (Assumption E) in Section 3.

 In addition, by \eqref{nonpaexpre}, we see that the nonparametric solution has a closed form in terms of the parametric coefficients, and the distributed estimation of the nonparametric components in \eqref{submethod} can could also be formulated. In general, the parametric estimation in PLM is more difficult than the nonparametric part, requiring more refined theoretical analysis. 

\section{Theoretical Results}
In this section, we first present several assumptions used in our theoretical analysis and introduce further notations. Thereafter, the assumptions are explained explicitly and the main results are stated.

{\bf Assumption A } (Model Specification).
For partially linear model \eqref{gmodel},  we assume that (i) $\bbeta^*$  is sparse with sparsity
$s^*$, and the nonparametric component $f^*$ is a multivariate zero-mean function in the RKHS defined pn $\mathcal{T}$; (ii) the noise terms $\epsilon_i$'s are independent normal variables, and also uncorrelated with covariates $(X_i,T_i)$.

{\bf Assumption B} (Covariates Behaviors). (i) The covariate $X$ in $\RR^p$ is bounded uniformly, that is, $|X_{\cdot j}|\leq C_0$ for all  $j\in[p]$.
(ii) The largest eigenvalue of the covariance matrix $\Sigma=\mathbb{E}[XX']$ is finite, denoted by $\Gamma_{\max}$.

Gaussian error assumption is a quite strong, but standard in the literature. In general, this condition can be easily relaxed to sub-Gaussian errors. It is worth noting that we allow correlations between $X$ and
$T$. The assumption that the target function belongs to the RKHS is frequently used in machine learning and statistics literature, see \citet*{Steinwart2008,Raskutti2012, Zhao2016} and many others.
A bound on the $X$-values is a more restrictive assumption than the sub-Gaussian tails, and we use it for technical reasons.

%In probability, concentration results with the sub-Gaussian random variables have been established well \citep{Pinelis1994,Vershynin2011}, and it suffices to conduct our analysis. 

%Throughout this paper, we assume that the nonparametric minimizer $\hat{f}$  in \eqref{submethod} can be found in the bounded ball $\mathcal{H}_R=\{f,\,\|f\|_K\leq C_K\}$ with some constant $C_K$. This assumption is standard in the literature of nonparametric estimation in RKHS \citep{Raskutti2012}.
To estimate the parametric and nonparametric parts respectively, 
we need some conditions concerning correlations between $X$ and $T$.
For each $j\in[p]$, let $\Pi^{(j)}_T$ be the projection of $X^{(j)}$ onto $\mathcal{H}_K$. That is, $\Pi^{(j)}_T=g_j^*(T)$ with
$$
g_j^*=\arg\min_{g\in\mathcal{H}_K}\mathbb{E}_{X^{(j)},T}[(X^{(j)}-g(T))^2].$$
We write $\Pi_{X|T}=(\Pi^{(1)}_T,...,\Pi^{(p)}_T)'$ and $X_T=X-\Pi_{X|T}$. Each function $g^*_j$ can be viewed as the best approximation  of $\mathbb{E}[X^{(j)}|T]$ within  $\mathcal{H}_K$.
In the extreme case, when $X$ is uncorrelated with $T$,
we get $\Pi_{X|T}=0$. The following condition ensures that there is enough information in the data to identify the parametric coefficients,
which has been imposed in \cite{Yu2011,Muller2015}.

{\bf Assumption C} (Mutual Correlation). (i) The smallest eigenvalue $\Lambda_{\min}$ of
$\mathbb{E}[X_TX_T']$ is positive.
% (2) $|(X_{\pi})_{ij}|\leq c_3$ for all $i\in[n]$ and $j\in[p]$. 
(ii) The largest eigenvalue of $\Sigma_\pi=\big(\langle g_k^*,g_l^* \rangle_K\big)_{k,l=1,...,p} $ is finite with high probability, denoted by $\Lambda_{K}$. 

%By the definition of projection and the reproducing property of RKHS, we see that $\|\Pi^{(j)}_T\|_\infty\leq C_K$ for any $j\in[p]$.

{\bf Assumption D} ( Spectral and super-norm assumption). (i): For some $0<\tau<1$, there exist two universal constants $C_1,C_2>0$, such that
$$
\mu_\ell\leq C_1\ell^{-1/\tau}.
$$
(ii) Additionally, there also holds
$$
\|g\|_\infty\leq C_2\|g\|_2^{1-\tau}\|g\|_{K}^{\tau},\quad \forall\, g\in \mathcal{H}_K.
$$
%Under mild conditions on $\mathcal{H}_K$, Assumption E is equivalent to the spectral decay, as stated in Assumption C. See the related details in  \cite{Steinwart2009}. The introduction of Assumption D is 
This assumption is satisfied if the RKHS is a
Sobolev space or is continuously embeddable in a Sobolev space. For instance,
the RKHSs of Gaussian kernels are continuously embedded in all Sobolev spaces, and thus satisfy sup-norm assumption. More general sufficient conditions to guarantee Assumption D(ii) are related to {\it real interpolation} shown in  \cite{Steinwart2009}.

{\bf Assumption E} (Generalized Coherence). Given $\tilde{\Sigma}_l$ defined on the $l$-the subsample,  the generalized coherence between $\tilde{\Sigma}_l$  and $\Xi \in R^{p\times p}$
$$
GC(\tilde{\Sigma}_l,\Xi)=\max_{j\in[p]}\|\tilde{\Sigma}_l\Xi'_j-\be_j\|_\infty.
$$
We assume that
$$
GC(\tilde{\Sigma}_l,\Xi)=O_p(\sqrt{\log p/n}),\quad \max_{j\in[p]}\big\|\Xi^{-1}_j-\hat{\Theta}^{(l)}_j\big\|_1=O_p(s_{max}\sqrt{\log p/n}),
$$
where $s_{\max}$ is the largest sparsity of the rows of $\Xi^{-1}$.

The preceding definition is viewed as a generalization of the standard coherence  appearing in the compressed sensing literature. It is worth noting that, $\tilde{\XX}^{(l)}$ defined as above is viewed as an adjustment of $\mathbb{X}$ for dependence on $T$. Hence, \cite{Javanmard2014} and Lemma 2 in \cite{Lee2015} for pure linear models is  also applicable  to our semi-parametric setting given any $T$, that is,
the GC condition is fulfilled when $X-\mathbb{E}[X|T]$ for almost $T$ are subgaussian random vectors and $\Xi$ is strictly positive. Besides, Theorem 2.4 in \cite{Geer2014} shows that $\hat{\Theta}^{(l)}_j$ converges to $\Xi^{-1}_j$ 
at the usual convergence rate of the lasso.

We are now ready to state our main results concerning the debiased averaging estimator \eqref{avergingestim}. It indicates that the averaging estimator achieves the convergence rate of the centralized double-regularized estimator in some cases,  as long as the dataset is not split across too many machines.

\begin{thm}\label{globalesmator}
	Suppose that Assumptions A-E hold.  we consider the debiased averaging parametric estimator defined by \eqref{avergingestim}, with suitable parameters constraint: 
	$$
\lambda^0\simeq\lambda_1\simeq \sqrt{\log p/n},\quad \lambda_2\simeq n^{-\frac{1}{\tau+1}}.
	$$
Let
	  $\lambda^0:=\lambda_1^{(j)}=...=\lambda_1^{(p)}$ in \eqref{covsolver}, we have 
	$$
	\|\bar{\bbeta}-\bbeta^*\|_\infty=O_p\Big((s^*+s_{max})
	\frac{\log p}{n}+\sqrt{\lambda_{\max}(\hat{\Theta}^{(l)})}n^{-\frac{(2+\tau)}{2(1+\tau)}}+\sqrt{\max_j\{(\hat{\Theta}^{(l)})_{jj}\}}\sqrt{\frac{\log p}{N}}\Big).
	$$ 
%	with probability at least 
%\begin{eqnarray*}
%	1-
%\end{eqnarray*}
% In particular,
%	with the choices of $ \lambda_1=$, $\lambda_2=\log p/N$ and $\lambda^0\simeq r_n=\sqrt{\log p/n}$, 
% we can choose the number of machines by $m\leq \sqrt{N/\log p}$, such that
%	$$
%	\|\bar{\bbeta}-\bbeta^*\|_\infty=
%	$$
\end{thm}
In the following, we assume that $s^*\simeq s_{max}$ for notional simplicity.
The rate of Theorem \ref{globalesmator} may be interpreted as the sum of estimation error of the parametric part
$O\big(s^*\frac{\log p}{n}\big)$, and the influence of the nonparametric component in PLM $O\big(n^{-\frac{(2+\tau)}{2(1+\tau)}}\big)$, as well as the total noise error
$O\big(\sqrt{\frac{\log p}{N}}\big)$. When $p$ is exponential of $n$ (e.g. $\log p=O(n^r)$ with $\frac{\tau}{2(1+\tau)}<r<1$),  $\big(n^{-\frac{(2+\tau)}{2(1+\tau)}}\big)$ is negligible compared to  $\big(s^*\frac{\log p}{n}\big)$ and the error of $\|\bar{\bbeta}-\bbeta^*\|_\infty$ is of the order $s^*\big(\sqrt{\frac{\log p}{N}}\big)$ by choosing $m\leq \sqrt{N/\log p}$. 
Otherwise, if the term $\big(n^{-\frac{(2+\tau)}{2(1+\tau)}}\big)$ dominates the parametric error, the total error $s^*\big(\sqrt{\frac{\log p}{N}}\big)$ can be also achieved by choosing $m\leq N^{\frac{1}{2+\tau}}/((\log p)^{\frac{1+\tau}{2+\tau}})$.
Up to the logarithmic term, the smallest number of data partition (corresponding to the worst case $\tau=1$) is $m=N^{\frac{1}{3}}$. It is interesting to note that the number of splits may be  larger as the nonparametric function becomes smoother. In summary, the partition-based parametric estimator achieves the statistical minimax rate over all estimators using the set of $N$ samples. 
%Precisely, the more complex the nonparametric component is, the smaller data partition is required
 
 %shows that the averaging estimator achieves the statistical minimax rate over all estimators using the set of $N$ samples, as if there is no information loss induced by data partitioning. 
%The constraint $\{\bbeta, \|\bbeta-\bbeta^*\|_1\leq L_n\}$ has also been imposed in some existing works, for example in \citet{fanlv13,Yuan2017}. 
The result in Theorem \ref{globalesmator} also indicates that the choice of the number $m$ of subsampled datasets does not rely on
 $s^*$ , which means that $m$ is adaptive to the sparsity parameter-+.

%On the other hand, we can derive a rough upper bound of $L_n$ from the minimization problem \eqref{submethod} without contraint. However, as shown below, $L_n$ may be diverging as $n$ increases.

%\begin{cor}
%Suppose that Assumptions A-D hold and $|\beta_j|\leq R_0$ for all $j\in S$. Consider the double-regularized estimator $(\hat{\bbeta}, \hat{f})$ defined by \eqref{submethod},  such that
%$\lambda_1= ,\,\,
%\lambda^0=r_n=\sqrt{\log p/n},\,\lambda_2=\log p/N. 
%$
%When $m\leq \sqrt{N/\log p}$,  with probability at least $1-$, it holds that
%$$
%\|\bar{\bbeta}-\bbeta^*\|_\infty=.
%$$
%\end{cor}
As an illustration, for the RKHS with infinitely many eigenvalues which decay at a rate $\mu_\ell=(1/\ell)^{2\alpha}$ for some parameter $\alpha > 1/2$. This type of scaling covers the case of Sobolev spaces and Besov spaces. In this case we can check that $\nu_n=n^{-\frac{\alpha}{2\alpha+1}}$. 
\begin{cor}
Under the same conditions as in Theorem 1 within a Sobolev space with $\alpha$ derivatives. When $p$ diverges polynomially fast with  the local sample size $n$, by choosing  $\lambda^0\simeq\lambda_1\simeq \sqrt{\log p/n}$ and $\lambda_2\simeq n^{-\frac{2\alpha}{2\alpha+1}}$, the estimation error of the parametric estimator is bounded by
$$
\|\bar{\bbeta}-\bbeta^*\|_\infty=s^*O_p\big(\sqrt{\frac{\log p}{N}}\big),
$$
provided that the number of local machines $m$ satisfy the bound
$m\leq N^{\frac{2\alpha}{4\alpha+1}}$.
\end{cor}
Our above arguments implies the relation between the split number $m$ and the smooth parameter $\alpha$.
%When $p$ is less than
%exponential in $n$, $\nu_n$ dominates $\sqrt{\log p/n}$ and this implies that $\gamma_n=\nu_n$. Thus, we have $\|\bar{\bbeta}-\bbeta^*\|_\infty=s$  by choosing $m=\sqrt{N/\log p}$, which is suboptimal in the minimax sense unless $\alpha$ goes to infinity.
We remark that, the main challenge to deriving the optimal minimax rates for our averaging parametric estimator comes from the negative influence of the nonparametric component in PLM, and this differs from the  semi-parametric literature under the non-distributed setting, see (\cite{Chen1988,Zhu2017}) and so on.

\section{Estimation on  Local Estimators }
This section provides general upper bounds on $\|X_T'(\hat{\bbeta}-\bbeta^*)\|_2$ and $\|\hat{\bbeta}-\bbeta^*\|_1$ for the standard PLM \eqref{gmodel}. The novelty of our results lies in that optimal estimation errors for PLM are obtained by the zero-order optimization for the regularized method \eqref{submethod}, and particularly our novel technique is also applicable to various non-smooth learning problems, such as SVM and then quantile regression. 
We now state the sketch of our main proof. First, we provide a crucial inequality that characterizes the relation between the parametric estimator and the nonparametric estimator, see Theorem 2 below. Second, under the same conditions of Theorem 2, Proposition 1 shows that  our estimators are bounded uniformly in high probability.
Thus, our final results immediately follow from  the derived results. The proof idea we adopt is to avoid the use of the frist-order information for our scheme \eqref{submethod}, and importantly we fill up the gap in terms of techniques  in the semi-parametric literature.  
%to be in a broader range than typically assumed in the literature, and this requirement is essential to derive sharp error bounds of our averaging parametric estimator.
In this paper, we are particularly interested in estimating $\bbeta^*$ when it is sparse in diverging dimensions. 
%As a by-product, we also obtain  optimal nonparametric rates for each local estimate of \eqref{submethod} in the minimax sense.

\begin{thm}\label{linearthm}
	Suppose that Assumptions A-D hold, and we consider the double-regularized estimator $(\hat{\bbeta}, \hat{f})$ defined by \eqref{submethod},  with the following parameters constrains
\begin{equation*}
\lambda_1/2\geq r_n+2\big( 2N_0\sqrt{ (C_0+\Pi_{max}+2)/n}+(C_0+\Pi_{max})\sqrt{r_n/n}\big)\|\widehat{\Delta}_f\|_{K},\,\, \lambda_2 \Gamma_{\max}/\Lambda_{\min}\leq 1/2,
\end{equation*}
then for any $r_n>0$ we have
\begin{align}\label{localbound}
1/2\|X_T'\widehat{\Delta}_\beta\|_2^2+
\lambda_1\|\widehat{\Delta}_\beta\|_1
&\leq 3 \|\widehat{\Delta}_\beta\|_1^2r_n+\big(4\sqrt{s^*/\Lambda_{\min}}\lambda_1+2\lambda_2\|f^*\|_{K}\sqrt{\Gamma_{\max}/\Lambda_{\min}}\big)\|X_{T}'\widehat{\Delta}_\beta\|_2\nonumber\\
&	+
2\lambda_2\sqrt{\Lambda_{K}/\Lambda_{\min}}\|X_{T}'\widehat{\Delta}_\beta\|_2\|\widehat{\Delta}_f\|_{K},
\end{align}
	with probability at least 
\begin{eqnarray*}
	1-e\cdot p\exp\big(-\frac{cnr_n^2}{\max_{j\in[p]}C_{K,j}^2}\big)
	-2e\cdot p^2\exp\Big(-\frac{cnr_n^2}{4(C_0+\Pi_{max})^4}\Big)
	-2p\exp(-nr_n^2).
\end{eqnarray*}
Here  $\Pi_{max}=\max_{j\in[p]}\|g^*_j\|_\infty$.  $C_{j,K}$ and $N_0$ are two absolute constants appearing in the proof of Theorem 2 and Lemma \ref{talagrand} respectively.
\end{thm}
The proof of Theorem \ref{linearthm}, contained in Section 5, constitutes one main technical contribution of this paper. From the result of Theorem \ref{linearthm}, it is seen that the quadratic term $\|\widehat{\Delta}_\beta\|_1^2$ appears in the right hand of \eqref{localbound}. Hence, this result is useful only if  upper bounds of
$\|\widehat{\Delta}_\beta\|_1$ and $\|\widehat{\Delta}_f\|_{K}$ can be proved to be bounded uniformly in advance.
 Proposition 1 attempts at solving such problem.
% There is some related work assuming directly that $L_n$ is bounded, see \cite{fanlv13,Yuan2017}. This case easily yields the optimal oracle rates of the parametric estimator. In general, there is no restriction on $L_n$ in most of the literature. In the latter setting, 
Recall that, a standard technical proof for analyzing \eqref{localbound} is to first construct some event, and then prove that the desired results hold under the event, as well as that the event occurs with high probability, see \citet*{Muller2015} for details. However, their results are based on the use  of the first order optimization and thereby they cannot be directly extended to any non-smooth learning problem.  We also notice that, the  restricted strong convexity \citep*{Raskutti2012} and restricted eigenvalues constants \citep*{Bickel2009} are two key tools to derive sharp error bounds of the oracle results. These mentioned techniques can be used for the two-step estimation for PLM \citep*{Zhu2017}, nevertheless it seems quite difficult to apply  for our one-step approach. 

\begin{prop}\label{propbound}
Suppose that  Assumptions A-D hold, with $\lambda_1\simeq \sqrt{\log(2p)/n}$ and $\lambda_2\simeq n^{-\frac{1}{\tau+1}}$.
% with $R=\sqrt{\lambda_2}$.
Then, we have
$$
\|\hat\bbeta-\bbeta^*\|_1=\sqrt{s^*}O_p\Big(\frac{n^{\frac{1}{2}-\frac{1}{1+\tau}}}{\sqrt{\log p}}\Big),
$$
and
$$
\|\hat f -f^*\|_K=O_p(1). 
$$
\end{prop}
The results in Proposition \ref{propbound}  follow easily from Lemma \ref{boundedres} below with suitable choices of additional parameters.  Here we omit the proof details for Proposition 1.
Moreover, Lemma \ref{boundw} and Lemma \ref{Guassianprocess} in Appendix  guarantee that the event in  Lemma \ref{boundedres} holds in high probability tending to $1$.

Note that $0<\tau<1$ implies  $\|\hat\bbeta-\bbeta^*\|_1=O_p(1)$ provided that $s^*$ is rather small as compared to
$n$ and $p$. Thus, combining the results of Theorem 2 and Proposition 1, the following  theorem follows immediately from Theorem \ref{linearthm}.

\begin{thm}\label{paramestim}
With the same conditions of Theorem \ref{linearthm} with $s^*=O(n^{\frac{1}{1+\tau}-\frac{1}{2}})$. By choosing
$\lambda_1\simeq \sqrt{\log p/n}$ and $\lambda_2\simeq n^{-\frac{1}{\tau+1}}$, we have
$$
\|\hat\bbeta-\bbeta^*\|_2=O\Big(\sqrt{\frac{s^*\log p}{n}}\Big)
$$
and
$$
\|\hat\bbeta-\bbeta^*\|_1=O\Big(s^*\sqrt{\frac{\log p}{n}}\Big)
$$
with probability at least $1-c_1p^{-1}$ for some universal constant $c_1>0$. 
\end{thm}
 Remark that, our theoretical results regarding the properties of the parameter estimators in \eqref{submethod} are non-asymptotic.
Indeed, estimation error of the nonparametric estimator is also obtained from Lemma 1 below and the result of Theorem \ref{paramestim},
 but this is not our  current focus  and we omit the details.

We notice that these two rates are the same order as those standard Lasso for linear models, see \cite{Bickel2009}
and \cite{Geer2014}. In other words,  despite the presence of the non-parametric part, the parametric part can  be estimated with parametric rate under regularity conditions.
% However, the price to pay for estimating PLM is that
%the probablity that these rates hold is rather smaller as compared to the standard term 

\section{Proofs}
In this section, we provide the detailed proofs of Theorems 1-2.
%, as well as the bound $L_n$ based on the minimization problem in \eqref{submethod}.
% We present a detailed proof, deferring some useful lemmas to the appendices.
 First of all, we give the proof of each local parametric estimate, which is one of key ingredients for obtaining the oracle rates of the averaging estimator based on the entire data.  

\subsection{Quantitative Relation Between Local Estimators}
In this subsection, we focus on theoretical analysis on each local machine $l$ ($l\in[m]$) in \eqref{submethod}. For all the symbols and numbers, we drop their dependence on $l$
for notational simplicity. In what following, we write $\Delta_\beta=\bbeta-\bbeta^*$ and $\Delta_f=f-f^*$, and particularly
$\widehat{\Delta}_\beta=\hat{\bbeta}-\bbeta^*$ and $\widehat{\Delta}_f=\hat{f}-f^*$. 

{\bf  Proof of Theorem \ref{linearthm}}.
By the definition of $(\hat{\bbeta}, \hat{f})$ in \eqref{submethod}, we have $\mathcal{L}(\hat{\bbeta},\hat{f})\leq \mathcal{L}(\bbeta^*, \hat{f}+\Pi_{X|T}'\widehat{\Delta}_\beta)$, which means that
\begin{align*}
	&\frac{1}{2}\|\bY-X'\hat{\bbeta}-\hat{f}\|_n^2+
	\lambda_1\|\hat{\bbeta}\|_1+\frac{\lambda_2}{2} \|\hat{f}\|_{K}^2\\
	&\leq
	\frac{1}{2}\|\bY-X'\bbeta^*-(\hat{f}+\Pi_{X|T}'\widehat{\Delta}_\beta)\|_n^2+
	\lambda_1\|\bbeta^*\|_1+\frac{\lambda_2}{2} \|\hat{f}+\Pi_{X|T}'\widehat{\Delta}_\beta\|_{K}^2.
	\end{align*}
Since $\|\hat{\bbeta}\|_1=\|\hat{\bbeta}_S\|_1+\|\hat{\bbeta}_{S^c}\|_1$ and $\|\bbeta^*\|_1=\|\bbeta^*_S\|_1$ by sparsity assumption in model \eqref{gmodel},
by the triangle inequality, the last inequality   implies that
	\begin{align}\label{codecom}
	&\|X_T'\widehat{\Delta}_\beta\|_n^2+
	2\lambda_1\|\widehat{\Delta}_\beta\|_1\nonumber\\
	&\leq2\Big|\frac{1}{n}\sum_{i=1}^{n}\epsilon_i(X_{T})_i'\widehat{\Delta}_\beta\Big|+2\Big|\frac{1}{n}\sum_{i=1}^{n}\widehat{\Delta}_f(T_i)(X_{T})_i'\widehat{\Delta}_\beta\Big|
	+2\Big|\frac{1}{n}\sum_{i=1}^{n}(\Pi_{X|T})_i'(\widehat{\Delta}_\beta)(X_{T})_i'\widehat{\Delta}_\beta\Big|\nonumber\\
	&+4\lambda_1\|(\widehat{\Delta}_\beta)_S\|_1+\lambda_2 (\|\hat{f}+\Pi_{X|T}'\widehat{\Delta}_\beta\|_{K}^2- \|\hat{f}\|_{K}^2)\nonumber\\
	&:=2\Phi_1+2\Phi_2+2\Phi_3+4\Phi_4+\Phi_5.
	\end{align}
	We will give tight upper bounds for all the terms $\Phi_i$'s. 
	First, since $(X_i, T_i)$  are independent of $\epsilon_i$ and they are bounded and  Gaussian respectively,
	$\mathbb{E}[\epsilon_i(X_T)_{ij}]=0$ and $\epsilon_i(X_T)_{ij}$ is sub-Gaussian, whose  sub-Gaussian norms are upper
	bounded, denoted by $C_{K,j}$, which depends on the Gamma function and $C_0$ and $\|g_j^*\|_\infty$. 
	By the Hoffding-type inequality in Lemma \ref {suggaussianinq} and the union bounds, we have
	\begin{eqnarray}\label{Phi1}
	\Phi_1=\Big|\frac{1}{n}\sum_{i=1}^{n}\epsilon_i(X_{T})_i'\widehat{\Delta}_\beta\Big|\leq \max_j\Big|\frac{1}{n}\sum_{i=1}^{n}\epsilon_i(X_{T})_{ij}\Big|\|\widehat{\Delta}_\beta\|_1\leq r_n\|\widehat{\Delta}_\beta\|_1,
	\end{eqnarray}
	with probability at least $1-e\cdot p\exp\big(-\frac{cnr_n^2}{\max_{j\in[p]}C_{K,j}^2}\big)$.

	Next, consider the second term $\Phi_2=\Big|\frac{1}{n}\sum_{i=1}^{n}\widehat{\Delta}_f(T_i)(X_{T})_i'(\widehat{\Delta}_\beta)\Big|$. Note that
	$$
	\Phi_2\leq \max_j\Big|\frac{1}{n}\sum_{i=1}^{n}\widehat{\Delta}_f(T_i)(X_{T})_{ij}\Big|\|\widehat{\Delta}_\beta\|_1,
	$$
	and $\mathbb{E}[\widehat{\Delta}_fX_{T}^{(j)}]=0$ for each $j\in[p]$ by the definition of projection,  and thus the Talagrand's concentration inequality in Hilbert spaces in Lemma \ref{talagrand} below can be used directly.
	To apply for Talagrand's concentration inequality, and it suffices to bound $\mathbb{E}[\mathbf{Z}]$,
	$B$ and $U$  respectively, involved in  Lemma \ref{talagrand}. We shall claim that,
	with probability at least $1-2p\exp(-nr_n^2)$, we get
	\begin{eqnarray}\label{phi21}
	\Phi_2\leq 2\Big( 2N_0\sqrt{\frac{ C_0+\Pi_{max}+2}{n}}+(C_0+\Pi_{max})\sqrt{\frac{r_n}{n}}\Big)\|\widehat{\Delta}_f\|_{K}\|\widehat{\Delta}_\beta\|_1,
	\end{eqnarray}
	where $N_0$ appearing in Lemma \ref{talagrand} is some absolute constant, independent of $n$ or $p$. In fact,
	it is trivial if  $\widehat{\Delta}_f$ is zero, and thus it suffices to
	consider non-zero cases. To this end, we define the function set
	$$
	\mathcal{G}=\Big\{g(X,T)=\frac{X_T^{(j)} f(T)}{\|f\|_{K}},f\in \mathcal{H}_{K}-\{0\}\Big\},\,\,j\in[p]
	$$
	and let $\mathbf{Z}=\sup_{g\in \mathcal{G}}\big|\frac{1}{n}\sum_{i=1}^ng(X_i,T_i)\big|$. Here we often drop the dependence on $j$ for simplicity.  By the definition of projection, $\mathbb{E}[f(T)X_{T}^{(j)}]=0$ for any $f\in \mathcal{H}_{K}$, \eqref{rademacherbounds} can be applied to yield
	\begin{eqnarray}\label{zesp1}
	\mathbb{E}[\mathbf{Z}]=\mathbb{E}\Big[\sup_{g\in \mathcal{G}}\big|\frac{1}{n}\sum_{i=1}^ng(X_i,T_i)\big|\Big]\leq 
	2\mathrm{R}_n(\mathcal{G}_1),
	\end{eqnarray}
	where $R_1(\cdot)$ refers to the Rademacher complexity defined in Appendix and $\mathcal{G}_1=\big\{g(X,T)=X_T^{(j)} f(T),\,\,\|f\|_{K}= 1\big\}$.
	Note that,  $\|X_T^{(j)}\|_\infty\leq C_0+\|g^*_j\|_\infty$ for any $j$, by the contraction inequality and sub-additivity of Rademacher complexity, we easily obtain that
	$$
	\mathrm{R}_n(\mathcal{G}_1)\leq \mathrm{R}_n(X_T^{(j)} )+  \mathrm{R}_n(\mathbb{B}_{K}(1))\leq2\sqrt{\frac{C_0+\|g^*_j\|_\infty+2}{n}},
	$$
	where we used the conclusion in Lemma \ref{kernelradema}. It immediately follows from \eqref{zesp1} that
	\begin{equation}\label{uuperb}
	\mathbb{E}[\mathbf{Z}]\leq 4\sqrt{\frac{C_0+\|g^*_j\|_\infty+2}{n}}.
	\end{equation}
	In addition, by the definition of $\mathcal{G}$, we can take $U=C_0+\|g^*_j\|_\infty$ and $B=(C_0+\|g^*_j\|_\infty)^2$ in Lemma   \ref{talagrand}.
	As a consequence,  following \eqref{uuperb} and the derived quantities of $U,B$, we conclude from Lemma   \ref{talagrand} and the union bounds that
	\begin{equation}
	\mathbf{Z}\leq 4N_0\sqrt{\frac{ C_0+\Pi_{max}+2}{n}}+2(C_0+\Pi_{max})\sqrt{\frac{r_n}{n}}, 
	\end{equation}
	with probability at least $1-p\exp(-nr_n^2)$ with any $r_n\leq 1$.  Let $f=\widehat{\Delta}_f$, the above claim in \eqref{phi21} is justified. 
	
We now turn to  the term $\Phi_{3}$.
%  under the constraint $\|\widehat{\Delta}_\beta\|_1\leq L_n$. 
	Following the definition of	projection, $\mathbb{E}[\Pi^{(k)}_TX_{T}^{(j)}]=0$
	for all $j,k\in[p]$, then we have the following decomposition:
	\begin{align}\label{pihdecom}
	\Phi_{3}=\Big|\frac{1}{n}\sum_{i=1}^{n}(\Pi_{X|T})_i'\widehat{\Delta}_\beta(X_{T})_i'\widehat{\Delta}_\beta\Big|
	%\Big|(\widehat{\Delta}_\beta)'\Big(\frac{1}{n}\sum_{i=1}^{n}(\Pi_{X|T})_i(X_T)_i'\Big)\widehat{\Delta}_\beta\Big|\nonumber\\
%	&\leq\Big|\widehat{\Delta}_\beta'\Big(\frac{1}{n}\sum_{i=1}^{n}(X_{T})_i(X_{T})_i'-\mathbb{E}[X_{T}X_{T}']\Big)\widehat{\Delta}_\beta\Big|\nonumber\\
	=\Big|\widehat{\Delta}_\beta'\Big(\frac{1}{n}\sum_{i=1}^{n}(\Pi_{X|T})_i(X_T)_i'-\mathbb{E}[\Pi_{X|T}X_{T}']\Big)\widehat{\Delta}_\beta\Big|
	%\nonumber\\
%	&:=\Phi_{31}+\Phi_{32},
	\end{align}
%since $X_T=X-\Pi_{X|T}$.
	By a simple algebra, we have
	\begin{eqnarray*}\label{equvient}
	\Phi_{3}\leq  \|\widehat{\Delta}_\beta\|^2_1\max_{k,j\in[p]}\Big|\frac{1}{n}\sum_{i=1}^{n}\big[(\Pi_{X|T})_i(X_T)_i'-\mathbb{E}[\Pi_{X|T}X_{T}']\big]_{k,j}\Big|,
	\end{eqnarray*}
	where $(\Pi^{(j)}_{T}X_{T}^{(k)})_{i}-\mathbb{E}[(\Pi^{(j)}_{T})(X_{T}^{(k)})]$ are upper bounded  by $2\|g_j^*\|_\infty(C_0+\|g_j^*\|_\infty)$ by assumption. Applying
the Hoffding-type inequality in Lemma \ref{suggaussianinq} yields that 
$$
\Big|\frac{1}{n}\sum_{i=1}^{n}\big[(\Pi_{X|T})_i(X_T)_i'-\mathbb{E}[\Pi_{X|T}X_{T}']\big]_{k,j}\Big|\leq r_n,\quad \forall\,\, j,k\in[p],
$$
with probability at least $1-e\cdot p^2\exp\Big(-\frac{cnr_n^2}{4\Pi_{max}^2(C_0+\Pi_{max})^2}\Big)$ and here the union bounds are used again. Thus, we have
\begin{eqnarray}\label{equvien}
	\Phi_{3}\leq  \|\widehat{\Delta}_\beta\|_1^2r_n,
\end{eqnarray}
with the same  probability as above. 
%As the same arguments as above, we can also obtain that
%	\begin{equation}\label{Phi31}
%	\Phi_{32} \leq  L_nr_n\|\widehat{\Delta}_\beta\|_1,
%	\end{equation}
%	with probability at least $1-2p^2\exp\Big(-\frac{cnr_n^2}{C_{K}^2(C_0+C_K)^2}\Big)$.
%	Then, combing with \eqref{equvien} and \eqref{Phi31}, it follows form \eqref{pihdecom} that, 	with probability at least $1-2p^2\exp\Big(-\frac{cnr_n^2}{C_{K}^2(C_0+C_K)^2}\Big)
%	-2p^2\exp\Big(-\frac{cnr_n^2}{4C_{K}^4}\Big)$,
%	\begin{equation}\label{Phi3com}
%	\Phi_{3} \leq  2L_nr_n\|\widehat{\Delta}_\beta\|_1.
%	\end{equation}
On the other hand, a simple algebra shows that
\begin{eqnarray}\label{phi4}
\Phi_{4}=\lambda_1\|(\widehat{\Delta}_\beta)_S\|_1\leq 
\sqrt{s^*}\lambda_1\|\widehat{\Delta}_\beta\|_2
\leq\sqrt{s^*/\Lambda_{\min}}\lambda_1\|X_{T}'\widehat{\Delta}_\beta\|_2,
\end{eqnarray}
where the last inequality follows from  Assumption C.

It remains to consider the term $\Omega_{5}$. A direct computation yields that
\begin{align}\label{phi5}
\Phi_{5}&=\lambda_2 (\|\hat{f}+\Pi_{X|T}'\widehat{\Delta}_\beta\|_{K}^2- \|\hat{f}\|_{K}^2)\nonumber\\
&\leq \lambda_2 \big(\|\Pi_{X|T}'\widehat{\Delta}_\beta\|_{K}^2+2\|\Pi_{X|T}'\widehat{\Delta}_\beta\|_{K}\|\widehat{\Delta}_f\|_{K}+2\|f^*\|_{K}\|\Pi_{X|T}'\widehat{\Delta}_\beta\|_{K}\big)\nonumber\\
&\leq \lambda_2 \Lambda_{K}/\Lambda_{\min}\|X_{T}'\widehat{\Delta}_\beta\|_2^2+2\lambda_2\sqrt{\Lambda_{K}/\Lambda_{\min}}\|X_{T}'\widehat{\Delta}_\beta\|_2\|\widehat{\Delta}_f\|_{K}\nonumber\\
&+2\lambda_2\|f^*\|_{K}\sqrt{\Lambda_{K}/\Lambda_{\min}}\|X_{T}'\widehat{\Delta}_\beta\|_2
%&\leq \lambda_2 %\Lambda_{K}/\Lambda_{\min}\|X_{T}'\widehat{\Delta}_\beta\|_2^2+2\lambda_2\big(\sqrt{\Lambda_{K}/\Lambda_{\min}}C_K
%+\|f^*\|_{K}\sqrt{\Lambda_{K}/\Lambda_{\min}}\big)\|X_T'\widehat{\Delta}_\beta\|_2,\nonumber
\end{align}
where the  second inequality follows from Assumption B.	
	
In summary,  combining with \eqref{Phi1}, \eqref{phi21}, \eqref{equvien}, \eqref{phi4} and \eqref{phi5}, we conclude from \eqref{codecom} that  
	\begin{align}\label{keyinequ}
	&\|X_T'\widehat{\Delta}_\beta\|_n^2+
	2\lambda_1\|\widehat{\Delta}_\beta\|_1\nonumber\\
	&\leq 2\Big(r_n+2\big( 2N_0\sqrt{ (C_0+\Pi_{max}+2)/n}+(C_0+\Pi_{max})\sqrt{r_n/n}\big)\|\widehat{\Delta}_f\|_{K}\Big)\|\widehat{\Delta}_\beta\|_1\nonumber\\
	&+2 \|\widehat{\Delta}_\beta\|_1^2r_n+\big(4\sqrt{s^*/\Lambda_{\min}}\lambda_1+2\lambda_2\|f^*\|_{K}\sqrt{\Lambda_{K}/\Lambda_{\min}}\big)\|X_{T}'\widehat{\Delta}_\beta\|_2\nonumber\\
&	+
	\lambda_2 \Lambda_{K}/\Lambda_{\min}\|X_{T}'\widehat{\Delta}_\beta\|_2^2+
	2\lambda_2\sqrt{\Lambda_{K}/\Lambda_{\min}}\|X_{T}'\widehat{\Delta}_\beta\|_2\|\widehat{\Delta}_f\|_{K}
	\end{align}
	with probability at least 
	\begin{eqnarray*}
		1-e\cdot p\exp\big(-\frac{cnr_n^2}{\max_{j\in[p]}C_{K,j}^2}\big)
		-e\cdot p^2\exp\Big(-\frac{cnr_n^2}{4\Pi_{max}^2(C_0+\Pi_{max})^2}\Big)
		-2p\exp(-nr_n^2).
	\end{eqnarray*}
%Note that, we also used the equivalent relationship between $\|X_T'\widehat{\Delta}_\beta\|_2^2$ and $\|X_T'\widehat{\Delta}_\beta\|_n^2$ derived in \eqref{equvien} under the constraint of $\|\widehat{\Delta}_\beta\|_1\leq L_n$.

We now establish an equivalent  relationship between  $\|X_T'\widehat{\Delta}_\beta\|_n^2$ and $\|X_T'\widehat{\Delta}_\beta\|_2^2$ in high dimensions. To this end, a direct computation leads to
\begin{eqnarray*}\label{equvienthigh}
\big|\|X_T'\widehat{\Delta}_\beta\|_n^2-\|X_T'\widehat{\Delta}_\beta\|_2^2\big|\leq  \|\widehat{\Delta}_\beta\|^2_1\max_{k,j\in[p]}\Big|\frac{1}{n}\sum_{i=1}^{n}\big[(X_{T})_i(X_{T})_i'-\mathbb{E}[X_{T}X_{T}']\big]_{k,j}\Big|,
\end{eqnarray*}
where $(X_{T})_{ij}(X_{T})_{ik}-\mathbb{E}[(X_{T})_{ij}(X_{T})_{ik}]$ are upper bounded  by $2(C_0+\|g_j^*\|_\infty)^2$ by Assumption B. Applying
the Hoffding-type inequality in Lemma \ref{suggaussianinq} yields that 
$$
\Big|\frac{1}{n}\sum_{i=1}^{n}(X_{T})_{ij}(X_{T})_{ik}-\mathbb{E}[(X_{T})_{ij}(X_{T})_{ik}]\Big|\leq r_n,\quad \forall\,\, j,k\in[p],
$$
with probability at least $1-e\cdot \exp\Big(-\frac{cnr_n^2}{4(C_0+\|g_j^*\|_\infty)^4}\Big)$. Then, the union bounds implies that 
\begin{eqnarray}\label{equviente}
\big|\|X_T'\widehat{\Delta}_\beta\|_n^2-\|X_T'\widehat{\Delta}_\beta\|_2^2\big|\leq r_n\|\widehat{\Delta}_\beta\|^2_1,
\end{eqnarray}
with probability at least $1-e\cdot p^2 \exp\Big(-\frac{cnr_n^2}{4(C_0+\Pi_{max})^4}\Big)$.
Then, plugging \eqref{equvienthigh} into \eqref{keyinequ}, we obtain that
\begin{align}\label{keyinequte}
&\|X_T'\widehat{\Delta}_\beta\|_2^2+
2\lambda_1\|\widehat{\Delta}_\beta\|_1\nonumber\\
&\leq 2\Big(r_n+2\big( 2N_0\sqrt{ (C_0+\Pi_{max}+2)/n}+(C_0+\Pi_{max})\sqrt{r_n/n}\big)\|\widehat{\Delta}_f\|_{K}\Big)\|\widehat{\Delta}_\beta\|_1\nonumber\\
&+3 \|\widehat{\Delta}_\beta\|_1^2r_n+\big(4\sqrt{s^*/\Lambda_{\min}}\lambda_1+2\lambda_2\|f^*\|_{K}\sqrt{\Lambda_{K}/\Lambda_{\min}}\big)\|X_{T}'\widehat{\Delta}_\beta\|_2\nonumber\\
&	+
\lambda_2 \Lambda_{K}/\Lambda_{\min}\|X_{T}'\widehat{\Delta}_\beta\|_2^2+
2\lambda_2\sqrt{\Lambda_{K}/\Lambda_{\min}}\|X_{T}'\widehat{\Delta}_\beta\|_2\|\widehat{\Delta}_f\|_{K}
\end{align}
	with probability at least 
\begin{eqnarray*}
	1-e\cdot p\exp\big(-\frac{cnr_n^2}{\max_{j\in[p]}C_{K,j}^2}\big)
	-2e\cdot p^2\exp\Big(-\frac{cnr_n^2}{4(C_0+\Pi_{max})^4}\Big)
	-2p\exp(-nr_n^2).
\end{eqnarray*}
Based on this and with the choice of 
\begin{equation}\label{tobecheck}
	\lambda_1/2\geq r_n+2\big( 2N_0\sqrt{ (C_0+\Pi_{max}+2)/n}+(C_0+\Pi_{max})\sqrt{r_n/n}\big)\|\widehat{\Delta}_f\|_{K},\,\, \lambda_2 \Lambda_{K}/\Lambda_{\min}\leq 1/2,
	\end{equation}
then	with the same probability as above, we have
\begin{align}\label{keyinequtete}
1/2\|X_T'\widehat{\Delta}_\beta\|_2^2+
\lambda_1\|\widehat{\Delta}_\beta\|_1
&\leq 3 \|\widehat{\Delta}_\beta\|_1^2r_n+\big(4\sqrt{s^*/\Lambda_{\min}}\lambda_1+2\lambda_2\|f^*\|_{K}\sqrt{\Lambda_{K}/\Lambda_{\min}}\big)\|X_{T}'\widehat{\Delta}_\beta\|_2\nonumber\\
&	+
2\lambda_2\sqrt{\Lambda_{K}/\Lambda_{\min}}\|X_{T}'\widehat{\Delta}_\beta\|_2\|\widehat{\Delta}_f\|_{K}.
\end{align}
This completes the proof of Theorem 2. \hspace*{8cm}$\Box$

Remark that, given that the correlation between $X$ and $T$ is weak appropriately(e.g. Assumption C(i) holds),
the convergence of the parametric estimator $\hat \bbeta$ is guaranteed with suitable choices of $(\lambda_1,\lambda_2)$, as long as both $\|\widehat{\Delta}_f\|_{K}$ and $\|\widehat{\Delta}_\beta\|_1$ are bounded uniformly. In other words, the convergence of the non-parametric estimator $\hat f$ has no effect on the convergence of the parametric estimator.

Technically, an obvious difference from the existing related proof  lies
in the start point to the proof. In our analysis, $\mathcal{L}(\hat{\bbeta},\hat{f})\leq \mathcal{L}(\bbeta^*, \hat{f}+\Pi_{X|T}'\widehat{\Delta}_\beta)$ instead of the classical zero-optimization $\mathcal{L}(\hat{\bbeta},\hat{f})\leq \mathcal{L}(\bbeta^*, f^*)$ (see \cite{Muller2015}). Note also that, a lower bound of $\lambda_2$ is required to ensure that $\|\widehat{\Delta}_f\|_{K}$.

\subsection{Boundedness for Local Estimators}
This subsection is devoted to providing rough convergence rates for local estimators in \eqref{submethod}.
Obviously, this means that both $\|\widehat{\Delta}_f\|_{K}$ and $\|\widehat{\Delta}_\beta\|_1$ are bounded uniformly under suitable choices of $(\lambda_1,\lambda_2)$. The corresponding proof borrows  the techniques from
\citet*{Muller2015}. 
Let for each $R>0$ and write $\lambda=(\lambda_1,\lambda_2)$, and we define
$$
\Omega_{R,\lambda}(\bbeta,f):=\lambda_1\|\bbeta\|_1/(R\sqrt{\delta_0/2})+\sqrt{\|X'\bbeta+f(T)\|_2^2+\lambda_2\|f\|_K^2},\quad \bbeta\in \mathbb{R}^p, \,f\in \mathcal{H}_K,                                         $$
where $\delta_0$ is a fixed small constant, specified in the proof of Lemma \ref{boundedres} below. Define a constrained function set 
$$
\mathcal{G}(R):=\{g(Z)=X'\bbeta+f(T): \, \Omega_{R,\lambda}(\bbeta,f)\leq R\}.
$$
We now introduce two events related to empirical processes theory
$$
\mathcal{T}_1(\delta_0,R):=\Big\{(X,T),\,\sup_{g\in \mathcal{G}(R)}\big|\|g\|^2_n-\|g\|^2_2\big|\leq \delta_0 R^2\Big\}
$$ 
and
$$
\mathcal{T}_2(\delta_0,R):=\Big\{(X,T,\epsilon),\,\sup_{g\in \mathcal{G}(R)}\big|\bepsilon'g/n\big|\leq \delta_0 R^2\Big\}.
$$
With these notations,  define the event 
$$
\mathcal{T}(\delta_0,R):=\mathcal{T}_1(\delta_0,R)\cap \mathcal{T}_2(\delta_0,R).
$$
\begin{lem}\label{boundedres}
 Suppose that Assumptions A and C hold. If the regularization parameters $(\lambda_1,\lambda_2)$ satisfy 
$$
2\lambda^2_1s^*/\Lambda^2_{\min}\leq \delta_0R^2,\quad 2\lambda_2 \|f^*\|_{K}^2\leq \delta_0R^2,
$$
Then on $\mathcal{T}(\delta_0,R)$, there holds $\Omega_{R,\lambda}(\widehat{\Delta}_{\bbeta},\widehat{\Delta}_f)\leq R$.
\end{lem}

%{\bf Proof for  Lemma \ref{boundedres}.}
\begin{proof}
Let $s=R/(R+\Omega_{R,\lambda}(\widehat{\Delta}_{\bbeta},\widehat{\Delta}_f))$, and we define two intermediate components by 
$$\tilde{\bbeta}=s\hat \bbeta+ (1-s) \bbeta^*,\quad \tilde{f}=s\hat f+ (1-s) f^*.$$
Then we  have
$\tilde{g}(Z):=X'\tilde{\bbeta}+\tilde{f}(T)=s\hat g(Z)+ (1-s) g^*(Z)$, where $\hat g(Z)=X'\hat \bbeta+\hat f(T)$.
Note that
$$
\Omega_{R,\lambda}(\tilde\bbeta-\bbeta^*,\tilde{f}-f^*)=s\Omega_{R,\lambda}(\widehat{\Delta}_{\bbeta},\widehat{\Delta}_f)=
\frac{R \Omega_{R,\lambda}(\widehat{\Delta}_{\bbeta},\widehat{\Delta}_f)}{R+\Omega_{R,\lambda}(\widehat{\Delta}_{\bbeta},\widehat{\Delta}_f)}\leq R,
$$
which means that $\tilde g-g^*\in \mathcal{G}(R)$. To compete the proof, the above formulation tells us that, it suffices to further show that $\Omega_{R,\lambda}(\tilde\bbeta-\bbeta^*,\tilde{f}-f^*)\leq R/2$.

We now state the details of this claim. 
By convexity of our objectiveness \eqref{submethod} and 
 definition of $(\hat{\bbeta},\hat{f})$,  we have
 \begin{align*}
 &\frac{1}{2}\|Y-\tilde{g}\|_n^2+
 \lambda_1\|\tilde{\bbeta}\|_1+
 \frac{\lambda_2}{2} \|\tilde{f}\|_{K}^2\leq\nonumber\\
 &s\big(\frac{1}{2}\|Y-\hat{g}\|_n^2+
 \lambda_1\|\widehat{\bbeta}\|_1+
 \frac{\lambda_2}{2} \|\hat{f}\|_{K}^2\big)+(1-s)\big(\frac{1}{2}\|Y-g^*\|_n^2+
 \lambda_1\|\bbeta^*\|_1+
 \frac{\lambda_2}{2} \|f^*\|_{K}^2\big)\\
 &\leq \frac{1}{2}\|Y-g^*\|_n^2+
 \lambda_1\|\bbeta^*\|_1+
 \frac{\lambda_2}{2} \|f^*\|_{K}^2.
 \end{align*}
Using that $Y+g^*+\bepsilon$ in our model \eqref{gmodel}, the above inequality can be rewritten as 
\begin{align*}
\|\tilde{g}-g^*\|_n^2+2
\lambda_1\|\tilde{\bbeta}\|_1+
\lambda_2 \|\tilde{f}\|_{K}^2\leq 2\bepsilon'(\tilde{g}-g^*)/n+2\lambda_1\|\bbeta^*\|_1+
\lambda_2 \|f^*\|_{K}^2.
\end{align*}
As shown above, $\tilde g-g^*\in \mathcal{G}(R)$. Hence, on $\mathcal{T}(\delta_0,R)$, we have
\begin{align*}
\|\tilde{g}-g^*\|_2^2+2
\lambda_1\|\tilde{\bbeta}\|_1+
\lambda_2 \|\tilde{f}\|_{K}^2\leq 3\delta_0R^2+2\lambda_1\|\bbeta^*\|_1+
\lambda_2 \|f^*\|_{K}^2.
\end{align*}
Since $\|\tilde{\bbeta}\|_1=\|\tilde{\bbeta}_S\|_1+\|\tilde{\bbeta}_{S^c}\|_1$ and $\|\bbeta^*\|_1=\|\bbeta^*_S\|_1$ by sparsity assumption in model \eqref{gmodel},
the triangle inequality is applied to imply that
\begin{align}\label{ineq2}
&\|X'_T(\tilde{\bbeta}-\bbeta^*)\|_2^2+	\|\Pi_{X|T}'(\tilde{\bbeta}-\bbeta^*)+(\tilde{f}-f^*)\|_2^2+	
2\lambda_1\|\tilde{\bbeta}_{S^c}\|_1+\lambda_2 \|\tilde{f}\|_{K}^2\nonumber\\
&\leq 3\delta_0R^2+
2\lambda_1\|\tilde{\bbeta}_S-\bbeta^*_S\|_1+
\lambda_2 \|f^*\|_{K}^2.
\end{align}
Using $uv\leq u^2+v^2/4$ for any $u,v\in \mathbb{R}$, we get
\begin{align}\label{traing}
\lambda_1\|\tilde{\bbeta}_S-\bbeta^*_S\|_1&\leq \lambda_1\sqrt{s^*}\|\tilde{\bbeta}_S-\bbeta^*_S\|_2\leq \lambda_1\sqrt{s^*}\|\tilde{\bbeta}-\bbeta^*\|_2
\leq \frac{\lambda_1\sqrt{s^*}}{\Lambda_{\min}}\|X'_T(\tilde{\bbeta}-\bbeta^*)\|_2\nonumber\\
&\leq  \frac{\lambda^2_1s^*}{\Lambda^2_{\min}}+\frac{\|X'_T(\tilde{\bbeta}-\bbeta^*)\|_2^2}{4}.
\end{align}
Meanwhile, observing that $\|\tilde{f}\|_{K}^2\geq \frac{1}{2}\|\tilde f-f^*\|_{K}^2-\|f^*\|_{K}^2$, we conclude from
\eqref{ineq2} that
\begin{align*}
&\|X'_T(\tilde{\bbeta}-\bbeta^*)\|_2^2+	\|\Pi_{X|T}'(\tilde{\bbeta}-\bbeta^*)+(\tilde{f}-f^*)\|_2^2+	
2\lambda_1\|\tilde{\bbeta}_{S^c}\|_1+\lambda_2/2 \|\tilde{f}-f^*\|_{K}^2\nonumber\\
&\leq 3\delta_0R^2+\frac{2\lambda^2_1s^*}{\Lambda^2_{\min}}+\frac{\|X'_T(\tilde{\bbeta}-\bbeta^*)\|_2^2}{2}
+
2\lambda_2 \|f^*\|_{K}^2.
\end{align*}
By our assumptions 
$$\frac{2\lambda^2_1s^*}{\Lambda^2_{\min}}\leq \delta_0R^2,\quad 2\lambda_2 \|f^*\|_{K}^2\leq \delta_0R^2,
$$
we obtain
\begin{align}\label{ineq3}
&1/2\|X'_T(\tilde{\bbeta}-\bbeta^*)\|_2^2+	\|\Pi_{X|T}'(\tilde{\bbeta}-\bbeta^*)+(\tilde{f}-f^*)\|_2^2+	
\lambda_2/2 \|\tilde{f}-f^*\|_{K}^2\nonumber\\
&\leq 5\delta_0R^2.
\end{align}
Hence, by  orthogonal decomposition of $\Pi_{X|T}$, this leads to
\begin{align*}
\|\tilde{g}-g^*\|_2^2+	
\lambda_2 \|\tilde{f}-f^*\|_{K}^2
\leq 10\delta_0R^2,
\end{align*}
that is,
\begin{align}\label{ineqexnted}
\big(\|X'(\tilde{\bbeta}-\bbeta^*)+(\tilde{f}-f^*)\|_2^2+	
\lambda_2 \|\tilde{f}-f^*\|_{K}^2\big)^{1/2}
\leq (10\delta_0)^{1/2}R.
\end{align}
On the other hand, note also that $\|\hat{\bbeta}_{S^c}\|_1=\|\hat{\bbeta}_{S^c}-\bbeta^*_{S^c}\|_1$ using sparsity assumption again, and adding $\|\hat{\bbeta}_S-\bbeta^*_S\|_1$ to both sides in \eqref{ineq2} yields that
\begin{align}\label{ineq32}
&\|X'_T(\tilde{\bbeta}-\bbeta^*)\|_2^2+	\|\Pi_{X|T}'(\tilde{\bbeta}-\bbeta^*)+(\tilde{f}-f^*)\|_2^2+	
2\lambda_1\|\tilde{\bbeta}-\bbeta^*\|_1+\lambda_2 \|\tilde{f}\|_{K}^2\nonumber\\
&\leq 3\delta_0R^2+
4\lambda_1\|\tilde{\bbeta}_S-\bbeta^*_S\|_1+
\lambda_2 \|f^*\|_{K}^2\leq \frac{9}{2}\delta_0R^2+\frac{\|X'_T(\tilde{\bbeta}-\bbeta^*)\|_2^2}{4},
\end{align}
where Inequality \eqref{traing} and our assumptions \big($\frac{2\lambda^2_1s^*}{\Lambda^2_{\min}}\leq \delta_0R^2,\,\lambda_2 \|f^*\|_{K}^2\leq \delta_0R^2
$\big)  were applied again.  Therefore, we have
$$
\lambda_1\|\tilde{\bbeta}-\bbeta^*\|_1\leq \frac{9}{4}\delta_0R^2,
$$
and
\begin{align}\label{key2}
\frac{\lambda_1\|\tilde{\bbeta}-\bbeta^*\|_1}{R\sqrt{\delta_0/2}}\leq \frac{9}{4}\sqrt{\delta_0/2}R.
\end{align}
Thus, from \eqref{ineqexnted} and \eqref{key2} we have
$$
\Omega_{R,\lambda}(\tilde\bbeta-\bbeta^*,\tilde{f}-f^*)\leq (10\delta_0)^{1/2}R+\frac{9}{4}\sqrt{\delta_0/2}R
\leq \frac{R}{2},
$$
as long as the small constant $\delta_0$ satisfies $\sqrt{\delta_0}\big[\sqrt{10}+\frac{9\sqrt{2}}{8}\big]\leq \frac{1}{2}$. Finally, our desired result follows from the following equality:
$$
\Omega_{R,\lambda}(\tilde\bbeta-\bbeta^*,\tilde{f}-f^*)=s\Omega_{R,\lambda}(\widehat{\Delta}_{\bbeta},\widehat{\Delta}_f)=
\frac{R \Omega_{R,\lambda}(\widehat{\Delta}_{\bbeta},\widehat{\Delta}_f)}{R+\Omega_{R,\lambda}(\widehat{\Delta}_{\bbeta},\widehat{\Delta}_f)}.
$$
Thus, we derive the desired result in Lemma \ref{boundedres}.
\end{proof}
At this point, if the event $\mathcal{T}(\delta_0,R)$ holds in high probability, the result of lemma \ref{boundedres}
implies convergence of local estimator $(\hat \bbeta,\hat f)$ by choosing a proper small quantity of $R$.
However, this derived convergence rate is rough and precisely the rate of the parametric estimator  depends on that of the nonparametric estimator, which is suboptimal in the literature. In appendix, we will show that $\mathcal{T}(\delta_0,R)$ holds in high probability under appropriate choices of regularization parameters.

\subsection{Proof for Averaging Estimator}
To derive estimation error of our averaging parametric estimator,
we decompose the total error into three parts: the first part characterizes the estimation error of the local estimator and the error of inverse matrix approximation, the second part reflects the approximation error of the nonparametric components in  the RKHS, and the third part is referred to as total noise.

{\bf Proof for Theorem 1}. Recall that the averaged parametric estimator $\bar{\bbeta}$ defined on all the subsample
is given by
\begin{equation}\label{debasiedestim}
\bar{\bbeta}=\frac{1}{m}\sum_{l=1}^m \big[\hat{\bbeta}^{(l)}+\frac{1}{n}\hat{\Theta}^{(l)}(\XX^{(l)})'(\mathbb{I}-\mathbb{A}_u^{(l)}(\lambda_2))(\bY^{(l)}
-\XX^{(l)}\hat{\bbeta}^{(l)})\big],
\end{equation}
where $\hat{\bbeta}^{(l)}$ is any Lasso estimator generated by minimizing $\mathcal{Q}^{(l)}(\bbeta)$ in \eqref{lasso} on the $l$-th subsample.

First, substituting the partially linear model into \eqref{debasiedestim}, we get
\begin{align*}
	\bar{\bbeta}&=\frac{1}{m}\sum_{l=1}^m \big[\hat{\bbeta}^{(l)}-\frac{1}{n}\hat{\Theta}^{(l)}(\XX^{(l)})'(\mathbb{I}-\mathbb{A}_u^{(l)}(\lambda_2))\XX^{(l)}(
	\hat{\bbeta}^{(l)}-\bbeta^*)\big]\nonumber\\
	&+\frac{1}{m}\sum_{l=1}^m \big[\frac{1}{n}\hat{\Theta}^{(l)}(\XX^{(l)})'(\mathbb{I}-\mathbb{A}_u^{(l)})(\mathbf{f}^*)^{(l)}\big]
	+\frac{1}{m}\sum_{l=1}^m \big[\frac{1}{n}\hat{\Theta}^{(l)}(\XX^{(l)})'(\mathbb{I}-\mathbb{A}_u^{(l)})\bepsilon^{(l)}\big].
\end{align*}
Subtracting $\bbeta^*$ one both sides of the last inequality, we obtain
\begin{eqnarray}\label{subeuq}
\|\bar{\bbeta}-\bbeta^*\|_\infty\leq \Omega_1+\Omega_2 +\Omega_3,
\end{eqnarray}
where
$$
\Omega_1=\frac{1}{m}\sum_{l=1}^m\Big\|\big(\mathbb{I}-\frac{1}{n}\hat{\Theta}^{(l)}(\XX^{(l)})'
(\mathbb{I}-\mathbb{A}_u^{(l)}(\lambda_2))\XX^{(l)}\big)(\hat{\bbeta}^{(l)}-\bbeta^*)\Big\|_\infty
$$
and
$$
\Omega_2=\Big\|\frac{1}{m}\sum_{l=1}^m\frac{1}{n} \big[\hat{\Theta}^{(l)}(\XX^{(l)})'(\mathbb{I}-\mathbb{A}_u^{(l)})(\mathbf{f}^*)^{(l)}\big]\Big\|_\infty,
\Omega_3=\frac{1}{N}\Big\|\sum_{l=1}^m \big[\hat{\Theta}^{(l)}(\XX^{(l)})'(\mathbb{I}-\mathbb{A}_u^{(l)})\bepsilon^{(l)}\big]\Big\|_\infty.
$$
We first consider the   term $\Omega_1$. For any $l\in[m]$, it is straightforward to see each term in the sum is bounded by
\begin{align*}\label{omega}
&\big\|\big(\mathbb{I}-\frac{1}{n}\hat{\Theta}^{(l)}(\XX^{(l)})'
(\mathbb{I}-\mathbb{A}_u^{(l)}(\lambda_2))\XX^{(l)}\big)(\hat{\bbeta}^{(l)}-\bbeta^*)\big\|_\infty\\
&\leq \big\|\big(\mathbb{I}-\Xi^{-1}\tilde{\Sigma}_l\big)(\hat{\bbeta}^{(l)}-\bbeta^*)\big\|_\infty
+\big\|\big(\Xi^{-1}-\hat{\Theta}^{(l)}\big)(\hat{\bbeta}^{(l)}-\bbeta^*)\big\|_\infty
\\
&\leq \max_{j\in[p]}\big\|\Xi^{-1}_{j,.}\tilde{\Sigma}_l-\mathbf{e}_j\big\|_\infty\big\|\hat{\bbeta}^{(l)}-\bbeta^*\big\|_1+
\max_{j\in[p]}\big\|\big(\Xi^{-1}_j-\hat{\Theta}^{(l)}_j\big)\big\|_1\big\|\tilde{\Sigma}_l(\hat{\bbeta}^{(l)}-\bbeta^*\big)\|_\infty.
\end{align*}
By Assumption E, $\max_{j\in[p]}\big\|\Xi^{-1}_{j,.}\tilde{\Sigma}_l-\mathbf{e}_j\big\|_\infty=O_p(\sqrt{\log p/n})$ and $\max_{j\in[p]}\big\|\big(\Xi^{-1}_j-\hat{\Theta}^{(l)}_j\big)\big\|_1=O_p(s_{max}\sqrt{\log p/n})$.
By Theorem 3, we have $\big\|\hat{\bbeta}^{(l)}-\bbeta^*\big\|_1=O_p(s^*\sqrt{\log p/n})$. Also by the optimality conditions of the local estimators \eqref{lasso}, we get $\big\|\tilde{\Sigma}_l(\hat{\bbeta}^{(l)}-\bbeta^*\big)\|_\infty=O_p(\lambda_1)$. We put all the pieces together to obtain 
%By Theorem \ref{paramestim}, we have
%$$
%\big\|\hat{\bbeta}^{(l)}-\bbeta^*\big\|_1=O_p(s^*\lambda_1), \,\,\,\forall\,l\in[m].
%$$
%with high probability tending to $1$ as $n$ goes to infinity.
% This together with \eqref{omega} implies that
\begin{equation}\label{omega1}
\Omega_1= (s^*+s_{max})O_p\Big(\frac{\log p}{n}\Big),
\end{equation}
provided that all the conditions in Theorem \ref{linearthm} are satisfied.

Next, we provide an upper bound of $\Omega_2$ in \eqref{subeuq}, upper bounded uniformly by 
$\Big\|\frac{1}{n} \hat{\Theta}^{(l)}(\XX^{(l)})'(\mathbb{I}-\mathbb{A}_u^{(l)})(\mathbf{f}^*)^{(l)}\Big\|_{2}$ for all $l\in[m]$. We notice that
%\begin{align*}
%	\Big\|\frac{1}{n} \hat{\Theta}^{(l)}(\XX^{(l)})'(\mathbb{I}-\mathbb{A}^{(l)})(\mathbf{f}^*)^{(l)}\Big\|_{2}&\leq \frac{1}{n}\Big\|\hat{\Theta}^{(l)}(\XX^{(l)})'(\mathbb{I}-\mathbb{A}^{(l)})^{1/2}\Big\|_{2}
%\Big\|(\mathbb{I}-\mathbb{A}^{(l)})^{1/2}(\mathbf{f}^*)^{(l)}\Big\|_2\\
%	&\leq\frac{1}{\sqrt{n}}\|\hat{\Theta}^{(l)}\|_2^{1/2}
%	\Big\|(\mathbb{I}-\mathbb{A}^{(l)})^{1/2}(\mathbf{f}^*)^{(l)}\Big\|_2,
%\end{align*}
\begin{align}
\Big\|\frac{1}{n} \hat{\Theta}^{(l)}(\XX^{(l)})'(\mathbb{I}-\mathbb{A}_u^{(l)})(\mathbf{f}^*)^{(l)}\Big\|_{2}^2&=\frac{1}{n^2}\big((\mathbf{f}^*)^{(l)}\big)'(\mathbb{I}-\mathbb{A}_u^{(l)})\XX^{(l)}\hat{\Theta}^{(l)}
\hat{\Theta}^{(l)}(\XX^{(l)})'(\mathbb{I}-\mathbb{A}_u^{(l)})(\mathbf{f}^*)^{(l)}\nonumber\\
&=\frac{1}{n^2}tr\Big((\mathbb{I}-\mathbb{A}_u^{(l)})\XX^{(l)}\hat{\Theta}^{(l)}
\hat{\Theta}^{(l)}(\XX^{(l)})'(\mathbb{I}-\mathbb{A}_u^{(l)})(\mathbf{f}^*)^{(l)}\big((\mathbf{f}^*)^{(l)}\big)' \Big)\nonumber\\
&\leq \frac{1}{n^2}\lambda_{\max}\big((\mathbb{I}-\mathbb{A}_u^{(l)})\XX^{(l)}\hat{\Theta}^{(l)}
\hat{\Theta}^{(l)}(\XX^{(l)})'\big)tr\big((\mathbb{I}-\mathbb{A}_u^{(l)})(\mathbf{f}^*)^{(l)}\big((\mathbf{f}^*)^{(l)}\big)' \big)\nonumber\\
&\leq \frac{1}{n}\lambda_{\max}(\hat{\Theta}^{(l)})	\|(\mathbb{I}-\mathbb{A}_u^{(l)})^{1/2}(\mathbf{f}^*)^{(l)}\|_2^2,
\end{align}
where we used the second equality follows from $tr(AB)=tr(BA)$ and the first inequality is based on $tr(AB)\leq \lambda_{\max}(A)tr(B)$ for any $A\succeq0$ and $B$.
Next, we  define an map $S_t$ from $\mathcal{H}_K\rightarrow \mathbb{R}^n$ by $S_t(f)=(f(T_1),...,f(T_n))$. Denote by $S_t^*$ the adjoint map of
$S_t$, satisfying $\left\langle S_tf, V\right\rangle_2=\left\langle f, S_t^*V\right\rangle_K$ for any $f\in \mathcal{H}_K$ and $V\in \RR^n$.
% Since $f^*=L_K^{1/2}g^*$ with some $g^*\in \mathcal{H}_K$,
Thus, one gets
\begin{align*}
	\|(\mathbb{I}-\mathbb{A}_u^{(l)})^{1/2}(\mathbf{f}^*)^{(l)}\|_2^2&=
	\left\langle S_tf^*, (\mathbb{I}-\mathbb{A}_u^{(l)})S_tf^*\right\rangle_2=
	\left\langle f^*, S_t^*(\mathbb{I}-\mathbb{A}_u^{(l)})S_tf^*\right\rangle_K\\
%	&=	\left\langle L_K^{1/2}(g^*), S_t^*(\mathbb{I}-\mathbb{A}^{(l)})S_tL^{1/2}_K(g^*)\right\rangle_K\\
	&\leq \|S_t^*(\mathbb{I}-\mathbb{A}_u^{(l)})S_t\|_{op}\|f^*\|_K^2,
\end{align*}
where $\|\cdot\|_{op}$ refers to the operator norm on $\mathcal{H}_K$.
Moreover, by the  property of adjoint map in Lemma \ref{opertor}, we know that
\begin{align*}
	\|S_t^*(\mathbb{I}-\mathbb{A}_u^{(l)})S_t\|_{op}&=
\|(\mathbb{I}-\mathbb{A}_u^{(l)})^{1/2}S_tS_t^*(\mathbb{I}-\mathbb{A}_u^{(l)})^{1/2}\|_{op}\\
&=\|S_tS_t^*(\mathbb{I}-\mathbb{A}_u^{(l)})\|_{op}\\
&\leq \lambda_2,
%&\leq \|(\mathbb{I}-\mathbb{A}^{(l)})^{1/2}S_t\big(\frac{1}{n}S_t^*S_t-L_K\big)S_t^*(\mathbb{I}-\mathbb{A}^{(l)})^{1/2}\|_{op}\\
%&+\|(\mathbb{I}-\mathbb{A}^{(l)})^{1/2}S_t\big(\frac{1}{n}S_t^*S_t\big)S_t^*(\mathbb{I}-\mathbb{A}^{(l)})^{1/2}\|_{op}\\
%&\leq \sqrt{\frac{2}{n}}	\|S_t^*(\mathbb{I}-\mathbb{A}^{(l)})S_t\|_{op}+\lambda_{\max}(\mathbb{K}^{(l)}_t)\lambda_2\\
%&\leq \big[\sqrt{2n}+\lambda_{\max}(\mathbb{K}^{(l)}_t)\big]\lambda_2
%\lambda_{\max}(S_t^*(\mathbb{I}-\mathbb{A}^{(l)})S_t)\\
%	\|(\mathbb{I}-\mathbb{A}^{(l)})^{1/2}S_tS_t^*(\mathbb{I}-\mathbb{A}^{(l)})^{1/2}\|_2\\
%	&=\|(\mathbb{I}-\mathbb{A}^{(l)})^{1/2}\mathbb{K}^{(l)}_t(\mathbb{I}-\mathbb{A}^{(l)})^{1/2}\|_2\\
%&=\lambda_{\max}((\mathbb{I}-\mathbb{A}^{(l)})^{1/2}\mathbb{K}^{(l)}_t(\mathbb{I}-\mathbb{A}^{(l)})^{1/2})\\
%&=\lambda_{\max}(\mathbb{K}^{(l)}_t(\mathbb{I}-\mathbb{A}^{(l)}))\\
%	&\leq \max_{i\in[n]}\Big(\frac{\nu_i}{1+\nu_i/(n\lambda_2)}\Big)\\
%	&\leq ??,
\end{align*}
where  we used the fact $S_tS_t^*=\mathbb{K}^{(l)}_t$. % and the formula $Tr(AB)=Tr(BA)$ again.
Thus, this together with the above results immediately yields
\begin{equation}\label{omega2}
\Omega_2\leq \|f^*\|_K\sqrt{\lambda_{\max}(\hat{\Theta}^{(l)})}\sqrt{\frac{\lambda_2}{n}}.
\end{equation}

It remains to quantify $\Omega_3$.  Note that the $j$-th element of $\Omega_3$ has the form
$$
\Omega_{3j}=\frac{1}{N}\big|\sum_{l=1}^m\sum_{i=1}^nw_{ij}^{(l)}\epsilon^{(l)}_i\big|,\,\,j\in[p],
$$
where $w_{\cdot j}^{(l)}:=e_j^TW^{(l)}$ is denoted to be the $j$-th row of  $W^{(l)}:=\hat{\Theta}^{(l)}(\XX^{(l)})'(\mathbb{I}-\mathbb{A}_u^{(l)})$, $j\in[p]$.
Since  $\epsilon_i$ are independent on covariates $(X_i,T_i)$ by Assumption A and $\bepsilon^{(l)}$ for all $l\in[m]$ are not overlapping
by splitting sample independently, $\Omega_{3j}$ is the sum of zero-mean i.i.d. Gaussian random variables conditional on $(X_i,T_i)$, thereby
applying the Hoeffding-type inequality in Lemma \ref{suggaussianinq} implies a two-sides tail bound of the form
\begin{eqnarray}\label{omega3}
\mathbb{P}[|\Omega_{3j}|>t |\,(X,T)]\leq e.\exp\Big(-\frac{cN^2t^2}{2m\max_{l}\|w^{(l)}_{.j}\|_{2}^2}\Big),\,\,\,\forall \,t>0, \hbox{and for all}\,j\in[p].
\end{eqnarray}
We still need to provide an upper bound of $\max_{l,j}\|w^{(l)}_{.j}\|_{2}^2$ before completing  $\Omega_3$. In fact, a direct calculation yields that
\begin{align*}
	\|w^{(l)}_{.j}\|_{2}^2&=
	e_j^T \hat{\Theta}^{(l)}(\XX^{(l)})'(\mathbb{I}-\mathbb{A}_u^{(l)})^2
	\XX^{(l)}\hat{\Theta}^{(l)}e_j\\
	&\leq e_j^T \hat{\Theta}^{(l)}(\XX^{(l)})'(\mathbb{I}-\mathbb{A}_u^{(l)})
	\XX^{(l)}\hat{\Theta}^{(l)}e_j\\
	&\leq n(\hat{\Theta}^{(l)})_{jj}.
\end{align*}
Then, with probability at least $1-p^{-1}$, we conclude from \eqref{omega3} that
\begin{equation}\label{omega33}
\|\Omega_{3}\|_\infty\leq\sqrt{\max_j\{(\hat{\Theta}^{(l)})_{jj}\}}/c
\sqrt{\frac{\log(3p)}{N}}.
\end{equation}

Consequently, substituting  \eqref{omega1}, \eqref{omega2} and \eqref{omega33} into \eqref{subeuq}, we can complete the proof of Theorem \ref{globalesmator}. \hspace*{11cm}$\Box$

\section{Simulations}
We illustrate the performances of the distributed estimators via simulations. We generate the data from the model \eqref{gmodel}, where $\bbeta^*=(1,2,-1,0.5,-2,0,\ldots,0)$ and $\epsilon_i\sim N(0,4)$. We then generate a vector $Z_i$ in $\RR^p$ from a mean-zero multivariate Gaussian distribution with correlations ${\rm Cov}(Z_{ij}Z_{ij'})=0.3^{|j-j'|}$, $1\le j,j'\le p$ and then set $T_i=\Phi(Z_{i1})$ and $X_{ij}=Z_{ij},j=2,\ldots,p$, where $\Phi$ is the cumulative distribution function of the standard normal distribution so that $T_i\in (0,1)$. The nonparametric function is $f^*(t)=5\sin(2\pi t)/(2-\sin(2\pi t))$ and the RKHS is chosen to be the 3rd order Sobolev space.  We select the tuning parameters in the penalties by 5-fold cross-validation in each local machine.  %, and an independently generated hold-out data set is used to tune the threshold value after aggregation.

We compute the centralized estimator (CEN) for $\bbeta$, the naive aggregated estimator without using bias correction (NAI) and the proposed aggregated estimator after bias correction (ABC). The accuracy of the estimators is assessed by $\|\bbeta-\bbeta_0\|_\infty$. 

\begin{figure}
	\centering
	\includegraphics[width=10cm,height=8cm]{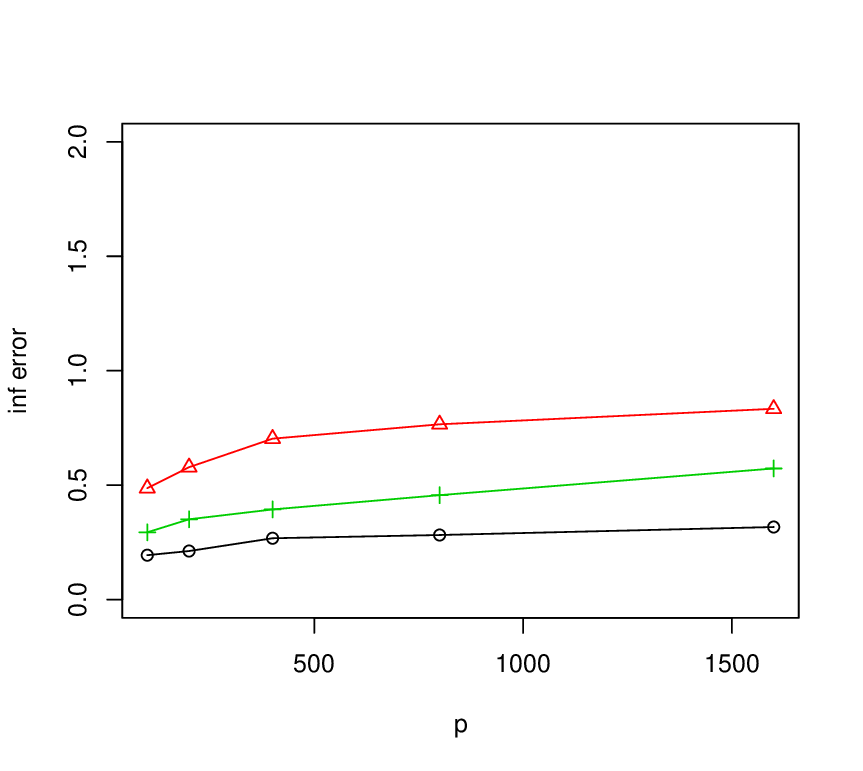} \ \
	\vspace{5mm}
	\caption{ The $l_\infty$ errors of estimates with changing dimension $p$. $\circ$(black): centralized estimator (CEN); $\triangle$(red): naive aggregated estimator (NAI); $+$(green): the aggregated estimator after debiasing (ABC).
		\label{fig:1} }
\end{figure}

First, we set $N=2000$, $m=1,10$ ($m=1$ is the centralized estimator) and $p=100,200,400,800,1600$. We generate $200$ data sets for each setting. Figure \ref{fig:1} shows the average errors of the centralized estimator (black) and those of the distributed estimators with $m=10$. It is seen the performance becomes worse with dimension as expected. The proposed aggregated estimator after bias correction (ABC) performs better than the naive aggregated estimator without using bias correction (NAI) for all dimensions.

In the second set of simulations, we vary $m=1,5,10,20,25$ while fixing $N=2000$ or $8000$, and $p=1000$. The performances generally deteriorate with the increase of $m$. Again, in terms of $l_\infty$ error, ABC is better than NAI. 

In the final set of simulations, we consider larger sample sizes $N=2000,4000,6000,\\8000,10000$, with $p=1000$, and fix the size of the sub-sample in each local machine to be $n=200$. It is seen that ABC has errors decreasing with total sample size, while the naive aggregated estimator NAI has larger errors.

The simulations are carried out on the computational cluster Katana in the Universityof New South Wales.  For the first set of simulations with $p=1600$ for example, the central estimators require about 8 hours  to  finish  all  200 repetitions, while the distributed estimator with $m=10$ requires about 1.5 hours.

\begin{figure}
	\centering
\begin{minipage}{0.5\textwidth}
  \centering
\includegraphics[width=0.8\textwidth]{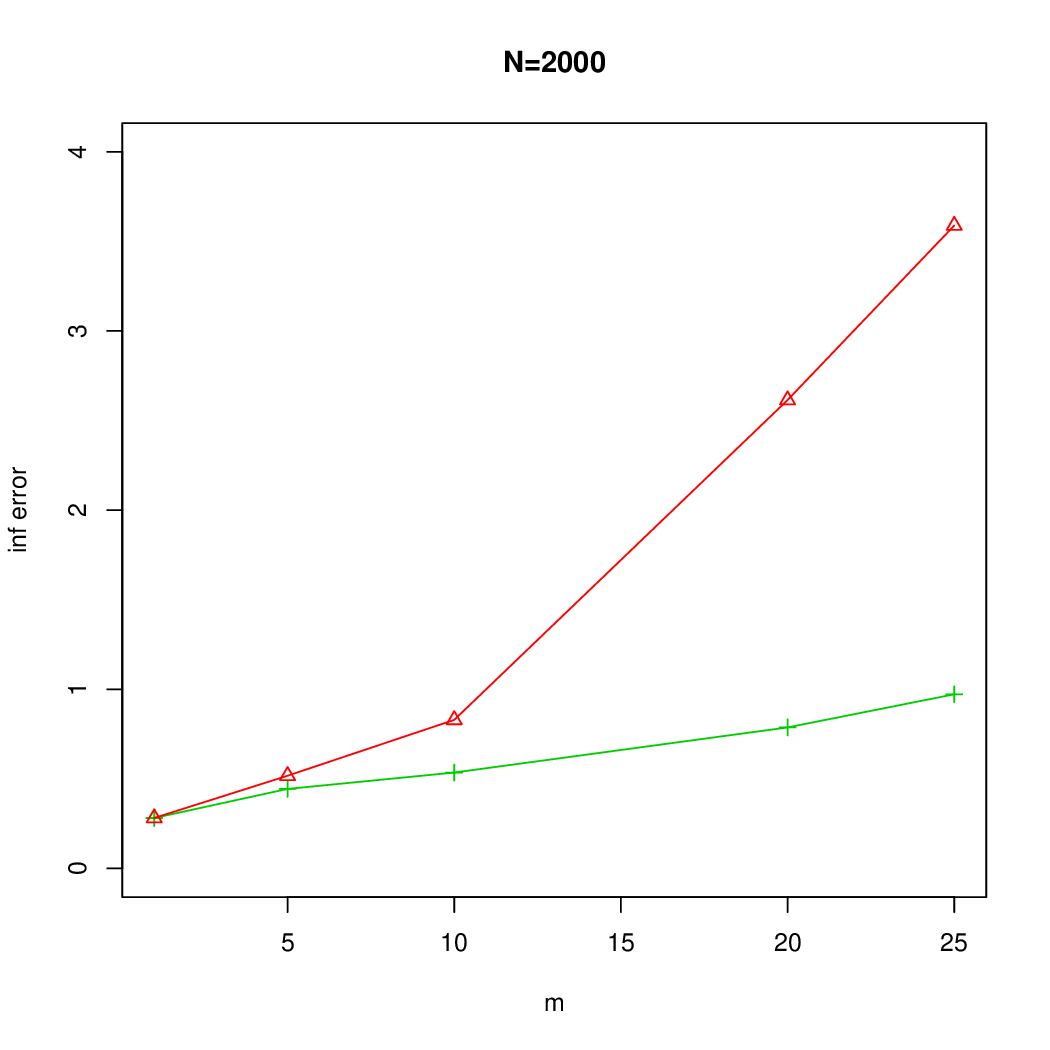}
%\subcaption[first caption.]{First}\label{fig:1a}
\end{minipage}%
\begin{minipage}{0.5\textwidth}
  \centering
\includegraphics[width=0.8\textwidth]{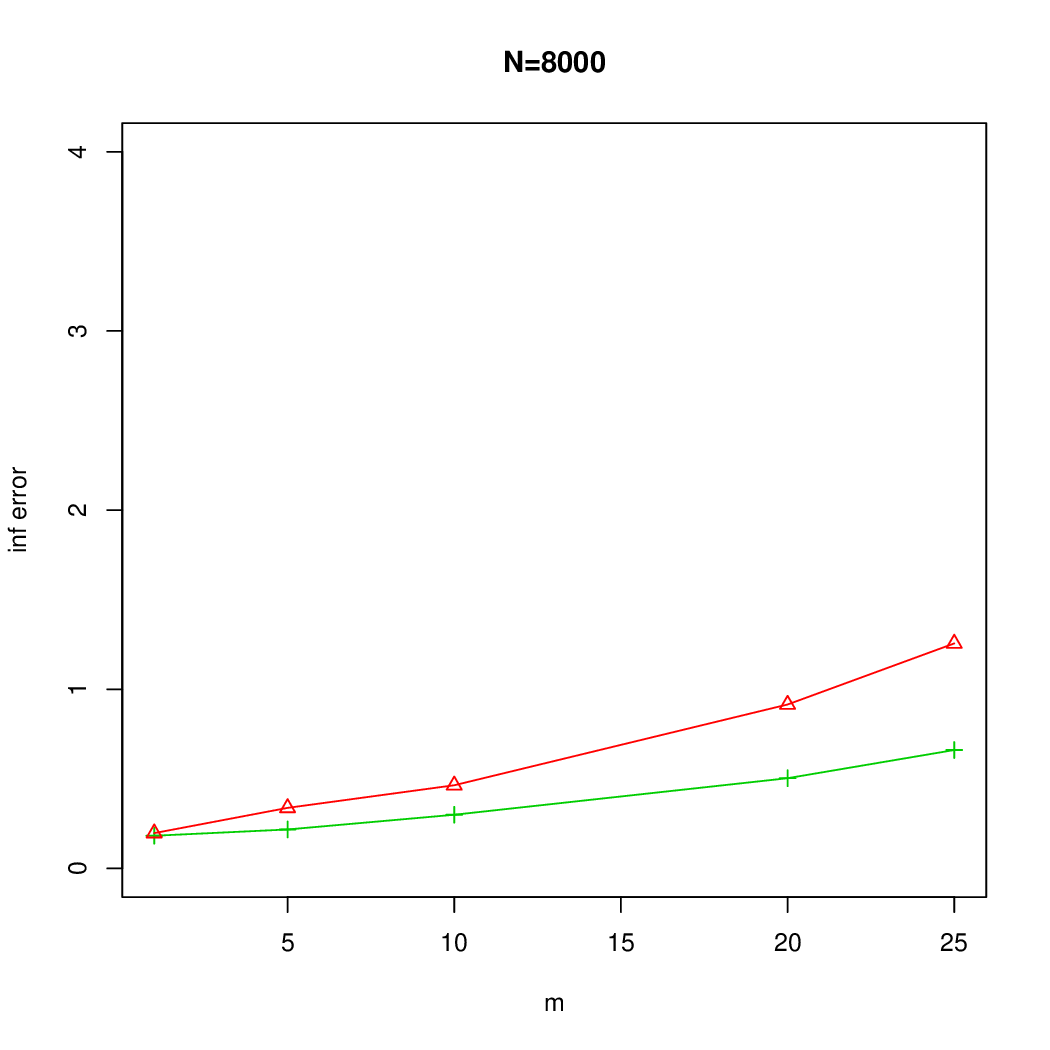}
%\subcaption[first caption.]{First}\label{fig:1a}
\end{minipage}%
	\vspace{5mm}
	\caption{ The $l_\infty$ errors of estimates with $m\in \{1,5,10,20,25\}$ ($m=1$ represents the centralized estimator).  $\triangle$(red): naive aggregated estimator (NAI); $+$(green): the aggregated estimator after debiasing (ABC). Left panel: $n=2000$, right panel: $n=8000$.
		\label{fig:2} }
\end{figure}

\begin{figure}
	\centering
	\includegraphics[width=10cm,height=8cm]{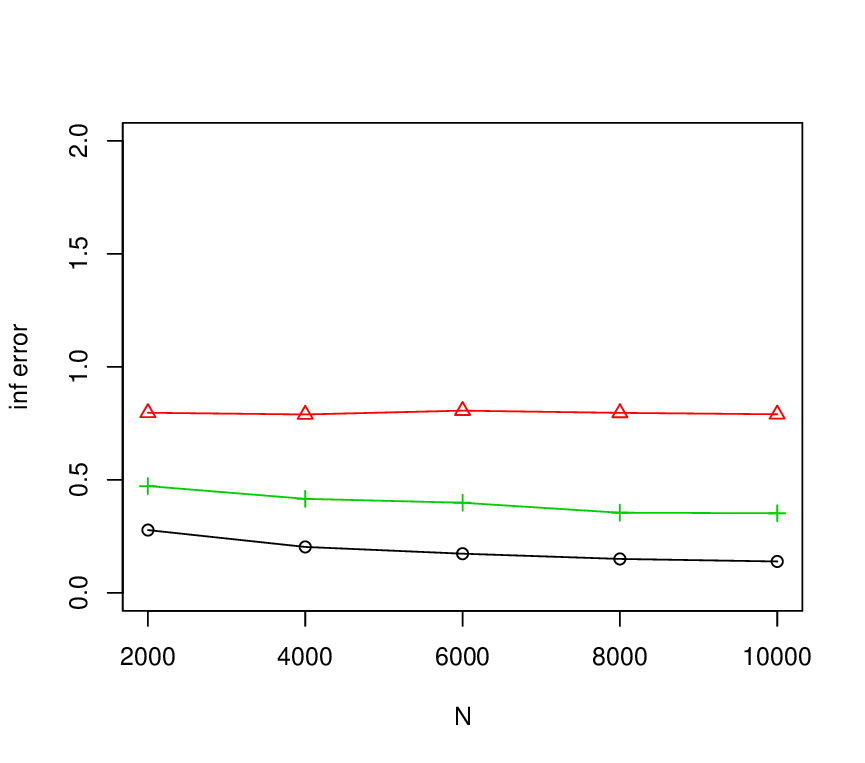} \ \
	\vspace{5mm}
	\caption{ The $l_\infty$ errors of estimates with $p=1000$ and $N\in \{2000, 4000, 6000, 8000, 10000\}$. $\circ$(black): centralized estimator (CEN);  $\triangle$(red): naive aggregated estimator (NAI); $+$(green): the aggregated estimator after debiasing (ABC).
		\label{fig:3} }
\end{figure}

\section{Conclusions}
Although distributed estimation or distributed learning have been studied well for linear models and fully nonparametric
models, to date  partial linear models  have been rarely studied under the distributed setting. The latter case encounters additional difficulty even in contrast to the centralized method on the entire data. As shown in the literature, the linear part in PLM can be estimated with oracle rates as if the nonparametric component were known, even though the rate for estimating the nonparametric component is slower than the oracle rate for the linear part. By contrast, to derive non-asymptotic oracle rates for the averaging parametric estimator,
the smoothness  of kernel-based nonparametric function significantly  affects the number of data partition. To handle this problem, we prove the oracle rate for the linear part with a novel technical proof, thereby yielding the minimax optimal rate of the parametric estimator in some senses. 

On the other hand, the classical double-regularized approach for estimating the sparse PLM heavily leads to estimation bias due to the two convex penalty terms. Hence, how to reduce bias is a critical issue to improve inference efficiency for the corresponding distributed estimation.  We transform the proposed estimation into a Lasso-type optimization only containing parametric coefficients, and then propose a new debiased distributed estimation for the sparse PLM under high dimensional setting, showing comparable numerical performance using several simulation experiments.

\bigskip
\bigskip

{\bf Acknowledgments}

We are grateful to two referees and the associate editor for valuable comments and constructive suggestions. The first author's research is supported partially by National Natural Science Foundation of China (Grant No.11871277 and 11829101).

\bigskip
\bigskip

\appendix
\section*{Appendix}

{\bf Appendix A. Concentration Inequalities and Complexity Bounds}

In this appendix we list several technical lemmas.

\begin{lem}(Talagrand's Concentration Inequality)\label{talagrand}
Let $\mathcal{G}$ be a function class on $\mathcal{Z}$ that is separable with respect to $\infty$-norm,
and $\{z_i\}_{i=1}^n$ be i.i.d. random variables with values in $\mathcal{Z}$. Furthermore, let $B>0$
and $U\geq0$ be $B:=\sup_{g\in \mathcal{G}}\mathbb{E}[(g-\mathbb{E}[g])^2]$ and
$U:=\sup_{g\in \mathcal{G}}\|g\|_\infty$, then there exists a universal constant $N_0$ such that, for
$\mathbf{Z}=\sup_{g\in \mathcal{G}}\big|\frac{1}{n}\sum_{i=1}^ng(z_i)-\mathbb{E}[g]\big|$, we have
$$
\mathbb{P}\Big(\mathbf{Z}\geq N_0\Big[\mathbb{E}[\mathbf{Z}]+\sqrt{\frac{Br}{n}}+\frac{Ur}{n}\Big]\Big)
\leq e^{-r},\quad \forall\,r>0.
$$
\end{lem}

We denote by $\{\sigma_i\}_{i=1}^n$ the Rademacher random variables that are an i.i.d. random variables taking values in $\{-1,+1\}$ with probability $1/2$.
Recall that, for a set of measurable functions $\mathcal{F}$ that is separable with respect to $\infty$-norm, the Rademacher complexity
$\mathrm{R}_n(\mathcal{F},\Phi(f)\leq r):=\mathbb{E}_{\sigma,u}\Big[\sup_{f\in \mathcal{F},\Phi(f)\leq r}\frac{1}{n}\big|\sum_{i=1}^n\sigma_if(u_i)\big|\Big]$ of $\mathcal{F}$ controls the supremum of discrepancy
between the empirical and population means of all functions $f\in \mathcal{F}$ (see Lemma 2.3.3 of van der Vaart and Wellner (1996)):
\begin{equation}\label{rademacherbounds}
\mathbb{E}\Big[\sup_{f\in \mathcal{F}}\big|\frac{1}{n}\sum_{i=1}^n(f(u_i)-\mathbb{E}[f])\big|\Big]\leq 2\mathrm{R}_n(\mathcal{F},\mathbb{E}[f^2]\leq r).
\end{equation}

\begin{lem}\label{contranin}
	Let $\mathcal{F}$ be a class of functions with ranges in $[a,b]$ and
there are some functional $\Phi: \mathcal{F}\rightarrow \RR^+$  such that for every $f\in \mathcal{F}$, $Var[f]\leq \Phi(f)\leq B\mathbb{E}[f]$. Let $\psi$ be a sub-root function and $r^*$ be the fixed point of $\psi$. Furthermore, assume that $\psi$ satisfies, for any
$r\geq r^*$,
$
\psi(r)\geq B \mathrm{R}_n(\mathcal{F},\Phi(f)\leq r).
$
Then, with $c_1=704$ and $c_2=26$, for any $L>1$ and every $r>0$, with probability at least $1-e^{-r}$,
$$
\frac{L}{L-1}\Big(\mathbb{E}[f]-\frac{1}{n}\sum_{i=1}^{n}f(u_i)\Big)\leq \frac{1}{L-1}\mathbb{E}[f]+\frac{c_1L}{B}r^*+\frac{r(11(b-a)+c_2LB)}{n},\,\,
\,\forall\,f\in \mathcal{F}.
$$ 
\end{lem}
The above concentration inequality can be viewed as a simple version of  Theorem 3.3 in \cite{Bartlett2005}.

The following result was proved in \cite{Mendelson2002}. There is an interesting finding that the upper bound of the Rademacher complexity in  the RKHS is independent of the dimension.  
\begin{lem}\label{kernelradema}
Suppose that the general kernel $K$ is  bounded uniformly by $\kappa$, then there holds
$$
\mathrm{R}_n(\mathbb{B}_K(1))\leq \sqrt{\frac{2\kappa}{n}}.
$$
Moreover, there also holds
$$
\mathrm{R}_n(f\in \mathbb{B}_K(1), \|f\|_2\leq r)\leq \mathcal{Q}_n(r).
$$
\end{lem}

For any $g=X'\bbeta+f(T)\in \mathcal{G}(R)$, we easily check that $\|f\|_K\leq \frac{R}{\sqrt{\lambda_2}}$. Moreover,
the traingle inequality is applied to obtain $\|g\|^2_2=\|X'_T\bbeta\|_2^2+	\|\Pi_{X|T}'\bbeta+f(T)\|_2^2\leq R^2$.
This together with Assumption C implies that $\|\bbeta\|_2\leq R/\Lambda_{\min}$. Furthermore, the triangle inequality is used to imply
$$
\|f\|_2\leq \|g\|_2+\|X'\bbeta\|_2\leq \big(\Gamma_{\max}/\Lambda_{\min}+1\big)R.
$$
Based on this, we obtain
\begin{align}\label{ramadercom}
\mathbb{E}\Big(\sup_{g\in \mathcal{G}(R)}\Big|\frac{1}{n}\sum_{i=1}^{n}f(T_i)\sigma_i\Big|\Big)\leq  \frac{R}{\sqrt{\lambda_2}}\mathrm{R}_n(f\in \mathbb{B}_K(1), \|f\|_2\leq \sqrt{\lambda_2})\leq 
\frac{\mathcal{Q}_n(\sqrt{\lambda_2})}{\sqrt{\lambda_2}}R.
\end{align}

%{\bf Some other useful Lemmas.}

The following lemma belongs to one of  large deviation inequalities for sums of independent sub-Gaussian random variables, and can be found in
Proposition 5.10 in \cite{Vershynin2011}. The sub-Guassian norm of $X$ is defined by
$\|X\|_{\psi_2}:=\sup_{p\geq 1}p^{-1/2}(\mathbb{E}|X|^p)^{1/p}$.
 \begin{lem}\label{suggaussianinq}(Hoffding-type inequality). Let $Z_1,...,Z_n$ be independent centered sub-gaussian random variables, and let $D=\max_i\|Z_i\|_{\psi_2}$. Then for every $a = (a_1,...,a_n)\in \mathbb{R}^n $ and every $t \geq 0$, we have
 $$
 \mathbb{P}\Big(\big|\sum_{i=1}^na_iZ_i\big|\geq t\Big)\leq e.
 \exp\Big(-\frac{ct^2}{D^2\|a\|_2^2}\Big),
 $$
 where $c$ is some universal constant.
\end{lem}

%\begin{lem}\label{subexp}
%Let $Z_1,...,Z_n$ be independent centered sub-exponential random variables, and let $C_{\psi_1} = \max_i\|Z_i\|_{\psi_1}$. Then, for every $r>0$, we have 
%$$
%\mathbb{P}\Big(\big|\frac{1}{n}\sum_{i=1}^nZ_i\big|\geq t\Big)\leq 2
%\exp\Big[-c\min\Big(\frac{r^2}{C_{\psi_1}^2},\frac{r}{C_{\psi_1}}\Big)n\Big],
%$$ 
%where $c>0$ is an absolute constant.
%\end{lem}

%The proof of Lemma \ref{boundfunc}  is based on the following concentration inequality in Hilbert spaces \citep*{Pinelis1994}.
%\begin{lem}\label{inftm}
%Let $\Xi_1$, ..., $\Xi_n$ be i.i.d. zero-mean random variables in a Hilbert space. If for some
%constants $M, V>0$, the bound $\mathbb{E}\|\Xi_i\|^\ell\leq
%\frac{1}{2}\ell!M^{\ell-2}V$ holds for every $2\leq\ell\in \mathbb{N}$, then there holds $$
%\mathbb{P}\Big(\big\|\frac{1}{n}\sum_{i=1}^n\Xi_i\big\|\geq r \Big)\leq 2 \exp\Big(-\frac{nr^2}{2(r M+V)}\Big), \qquad \forall\, r >0.$$
%\end{lem}

%\begin{lem}
%Given $\xi_i$, $1\leq i\leq n$ are i.i.d. random sub-Gaussian vectors
%with covariance matrix $\Sigma$
%\end{lem}

We introduce the event involving  $\nu_n$ defined in the text previously:
$$
\mathcal{E}(\nu_n)=\Big\{\big|\frac{1}{n}\sum_{i=1}^n\epsilon_i f(T_i)\big)\big|\leq  \nu_n^2 \|f\|_{K}+\nu_n\|f\|_{2},\quad f\in \mathcal{H}_{K}\Big\}.
$$
By simplifying Theorem 10 for multi-kernel regression problems in the supplementary material  of \cite{Suzuki2013}, one shows that the event $\mathcal{E}(\nu_n)$ occurs with
high probability, stated as follows.
\begin{lem}\label{weightemp}
	Suppose that $\epsilon_i$'s are independent Gaussian variables. Under the Supernorm Assumption, then there exist two
	constants $c_1,\,c_2$ such that
	$$
	\mathbb{P}[\mathcal{E}(\nu_n)]\geq 1-c_1\exp(-c_2n\nu_n^2).
	$$
\end{lem}

In the end, we list the classical conclusion on the adjoint operators in Hilbert spaces,
see the Chapter 8 in \cite{Rudin1991}. 
\begin{lem}\label{opertor}
	Let $H_1,H_2$ be two Hilbert spaces, and $A$ is a linear and bounded operator from $H_1$ to $H_2$, with its adjoint operator $A^*$. Then, $\|A\|=\|A^*\|=\|AA^*\|^{1/2}=\|A^*A\|^{1/2}$.  
\end{lem}

\bigskip

{\bf Appendix B. Proof for $\mathcal{T}_1(\delta_0,R)$}

%\begin{lem}\label{concentration}(Concentration Theorem \citep*{Bousquet2002})
%	Let $U_1,...,U_n$ be independent random variables with values in
%	some space $\mathcal{U}$ and let $H$ be a class of
%	real-valued functions on $\mathcal{U}$, satisfying for some
%	positive constants $\eta_n$ and $\tau_n$,
%	$$\|h\|_\infty\leq \eta_n,\quad \hbox{and}\,\,
%	\frac{1}{n}\sum_{i=1}^nvar(h(U_i))\leq \tau_n^2,\quad
%	\forall\,h\in H.$$ Define $\boldsymbol{Z}:=\sup_{h
%		\in
%		H}\Big|\frac{1}{n}\sum_{i=1}^n\big(h(U_i)-\mathbb{E}h(U_i)\big)\Big|.$
%	Then for $t>0$
%	$$\mathbb{P}\Big(\boldsymbol{Z}\geq \mathbb{E}(\boldsymbol{Z})+
%	t\sqrt{2(\tau_n^2+2\eta_n\mathbb{E}(\boldsymbol{Z}))}+
%	\frac{2\eta_nt^2}{3}\Big)\leq \exp[-nt^2].$$
%\end{lem}

\begin{lem}\label{boundw}
	Suppose that  Assumptions A-D hold, with $\lambda_2\simeq R^2$ and $ R^2\leq \lambda_1\leq 1$.
	For constants $\delta_1,\delta_1''$ and $\kappa_1$ with our  suitable choices, we set  $\lambda_0:=\sqrt{\log(2p)/n}$ and
	$$
	\delta_1\lambda_1\geq \lambda_0,\,\, \sqrt{n}\lambda_1\geq 1,\,\,\lambda_2\geq \kappa_1n^{-\frac{1}{\tau+1}}.
	$$
	Then we conclude 
	$$
	\sup_{g\in \mathcal{G}(R)}
	\big|\|g\|_n^2-\|g\|^2\big|\leq \delta_0 R^2
	$$
	with probability at least $1-\exp[-n(\delta_1'')^2\lambda_2]$.
\end{lem}

\begin{proof}
	In order to verify all the conditions of Lemma \ref{talagrand}, we denote $\boldsymbol{Z}:=\sup_{g\in \mathcal{G}(R)}\big|\|g\|_n^2-\|g\|^2\big|$ with
	$\mathcal{G}(R)$ and $Z:=(X,T)$. By a direct computation, we have
	$$
	\|g^2\|_\infty=\big\|\big(X'\bbeta+f(T)\big)^2\big\|_\infty\leq 2\|f\|_K^2+2C_0^2\|\bbeta\|^2_1.
	$$
	Note now that for $g\in  \mathcal{G}(R)$, it follows that
	$$
	\|\bbeta\|_1\leq \frac{\sqrt{\delta_0/2}R^2}{\lambda_1},\,\,\hbox{and}\,\,
	\|f\|_K^2\leq \frac{R^2}{\lambda_2},
	$$
	which  implies that
	\begin{align*}
	\|g^2\|_\infty\leq 2 \frac{R^2}{\lambda_2}+
	\delta_0C_0^2\frac{R^4}{\lambda_1^2}\leq  2/c(\delta_0) +
	\delta_0C_0^2,
	\end{align*}
	since $\lambda_2\geq c(\delta_0) R^2$ and that $R^2\leq \lambda_1\leq 1$ by assumption.
	Letting $\tilde{C}= 2/c(\delta_0) +\delta_0C_0^2$,
	for any $g\in \mathcal{G}(R)$, we further have
	\begin{eqnarray}
	var(g^2)\leq \mathbb{E}[g^4]\leq \|g^2\|_\infty\mathbb{E}[g^2]\leq \tilde{C} R^2.
	\end{eqnarray}
	We now need to provide an upper bound of $\mathbb{E}[\boldsymbol{Z}]$.
	Let $\{\sigma_i\}_{i=1}^n$ be a Rademacher sequence independent of 
	$\{(X_i,T_i)\}_{i=1}^n$. By symmetrization [see e.g. van der Vaart and Wellner (1996)], we have
	\begin{align*}
	\mathbb{E}[\boldsymbol{Z}]&\leq 2\mathbb{E}\Big(\sup_{g\in \mathcal{G}(R)}\Big|\frac{1}{n}\sum_{i=1}^{n}g_i^2\sigma_i\Big|\Big)
	\leq 2\mathbb{E}\Big(\sup_{g\in \mathcal{G}(R)}\Big|\frac{1}{n}\sum_{i=1}^{n}\langle f^2(T_i)\sigma_i\Big|\Big)\\
	&+
	2\mathbb{E}\Big(\sup_{g\in \mathcal{G}(R)}\Big|\frac{1}{n}\sum_{i=1}^{n}(X'_i\bbeta)^2\sigma_i\Big|\Big)+4\mathbb{E}\Big(\sup_{g\in \mathcal{G}(R)}\Big|\frac{1}{n}\sum_{i=1}^{n}(X'_i\bbeta)f(T_i)\sigma_i\Big|\Big).
	\end{align*}
	In the following, we bound the above three quantities respectively. Note that for
	$f(T)+X'\bbeta\in \mathcal{G}(R)$,  Condition B leads to
	$$
	\|X' \bbeta\|_\infty \leq C_0 \frac{\sqrt{\delta_0/2}R^2}{\lambda_1}\leq C_0 \sqrt{\frac{\delta_0}{2}},
	$$ 
	where we used the assumption that $R^2\leq \lambda_1$. By the contraction inequality of Rademacher complexity [see Ledoux and Talagrand (1991)], we get
	$$
	\mathbb{E}\Big(\sup_{\bbeta, f(T)+Z^T\bbeta \in \mathcal{G}(R)}\Big|\frac{1}{n}\sum_{i=1}^{n}(X_i'\bbeta)^2\sigma_i\Big|\Big)\leq 4 C_0 \sqrt{\frac{\delta_0}{2}} \mathbb{E}\Big(\sup_{\bbeta, f(T)+X'\bbeta \in \mathcal{G}(R)}\Big|\frac{1}{n}\sum_{i=1}^{n}(X'_i\bbeta)\sigma_i\Big|\Big).
	$$
	Moreover, we know that
	\begin{align*}
	&\mathbb{E}\Big(\sup_{\bbeta, f(T)+X'\bbeta \in \mathcal{G}(R)}\Big|\frac{1}{n}\sum_{i=1}^{n}(X_i'\bbeta)\sigma_i\Big|\Big)
	\leq\mathbb{E}\Big(\sup_{\bbeta, f(T)+Z^T\bbeta \in \mathcal{G}(R)} \Big\|\frac{1}{n}\sum_{i=1}^{n}X_i \sigma_i\Big\|_\infty\|\bbeta\|_1\Big)\nonumber\\
	&\leq \frac{\sqrt{\delta_0/2}R^2}{\lambda_1}\mathbb{E}\Big\|\frac{1}{n}\sum_{i=1}^{n}X_i \sigma_i\Big\|_\infty\leq C_0\frac{\sqrt{\log(2p)/n}R^2}{\sqrt{1/\delta_0}\lambda_1}
	\leq \big(\delta_1 C_0\sqrt{\delta_0}\big)R^2,
	\end{align*}
	where the first inequality follows from the Cauchy-Schwarz inequality,  the classical concentration result $\mathbb{E}\Big\|\frac{1}{n}\sum_{i=1}^{n}X_i \sigma_i\Big\|_\infty\leq C_0\sqrt{2\log(2p)/n}$ in the third one and the assumption
	$\delta_1\lambda_1\geq \sqrt{\log(2p)/n}$ for the last inequality. Hence, combining with the above two inequalities yields 
	\begin{align}\label{parradm}
	\mathbb{E}\Big(\sup_{\bbeta, f(T)+X'\bbeta \in \mathcal{G}(R)}\Big|\frac{1}{n}\sum_{i=1}^{n}(X_i'\bbeta)^2\sigma_i\Big|\Big)\leq 2\delta_1C_0^2\delta_0R^2.
	\end{align}
	
	At this point, we still require a tight bound on $\mathbb{E}\Big(\sup_{g\in \mathcal{G}(R)}\Big|\frac{1}{n}\sum_{i=1}^{n}f(T_i)^2\sigma_i\Big|\Big)$.
	As stated above, it is shown that
	$|f(T)|\leq \|f\|_K\leq  \frac{R}{\sqrt{\lambda_2}}$. By the contraction 
	property of Rademacher sequences again, we also have
	\begin{align}\label{squradem}
	\mathbb{E}\Big(\sup_{g\in \mathcal{G}(R)}\Big|\frac{1}{n}\sum_{i=1}^{n}f(T_i)^2\sigma_i\Big|\Big)
	&\leq 2\frac{R}{\sqrt{\lambda_2}}\cdot\mathbb{E}\Big(\sup_{g\in \mathcal{G}(R)}\Big|\frac{1}{n}\sum_{i=1}^{n}f(T_i)\sigma_i\Big|\Big)
	\nonumber\\
	&\leq 2\frac{\mathcal{Q}_n(\sqrt{\lambda_2})}{\lambda_2}R^2,
	\end{align}
	which follows from the obtained result in \eqref{ramadercom} in Appendix. Similarly, we also have
	\begin{align}\label{interam}
	&\hspace*{-1cm}\mathbb{E}\Big(\sup_{g\in \mathcal{G}(R)}\Big|\frac{1}{n}\sum_{i=1}^{n}(X_i'\bbeta)f(T_i)\sigma_i\Big|\Big)
	\leq \mathbb{E}\sup_{g\in \mathcal{G}(R)}\Big\|\frac{1}{n}\sum_{i=1}^{n}X_if(T_i)\sigma_i\Big\|_\infty\|\bbeta\|_1\\
	&\leq \sqrt{\delta_0/2}\frac{R^2}{\lambda_1}\mathbb{E}\max_{1\leq j\leq p}\sup_{f,X'\bbeta+f(T)\in \mathcal{G}(R)}\Big|\frac{1}{n}\sum_{i=1}^{n}X_{ij}f(T_i)\sigma_i\Big|\nonumber\\
	&\leq C_0\sqrt{\delta_0/2}\frac{R^2}{\lambda_1}\mathbb{E}\Big(\sup_{g\in \mathcal{G}(R)}\Big|\frac{1}{n}\sum_{i=1}^{n}f(T_i)\sigma_i\Big|\Big)\nonumber\\
	&\leq C_0\sqrt{\delta_0/2}\frac{\mathcal{Q}_n(\sqrt{\lambda_2})}{\lambda_1\sqrt{\lambda_2}}R\cdot R^2
	\end{align}
	where the third inequality follows from the contraction property of Rademacher complexity, and the last inequality follows from \eqref{ramadercom} below.
	Along the lines of \eqref{parradm}, \eqref{squradem} and \eqref{interam},
	we get
	$$
	\mathbb{E}[\boldsymbol{Z}]\leq \Big(2\delta_1C_0^2\delta_0+ 2\frac{\mathcal{Q}_n(\sqrt{\lambda_2})}{\lambda_2}+C_0\sqrt{\delta_0/2}\frac{\mathcal{Q}_n(\sqrt{\lambda_2})}{\lambda_1\sqrt{\lambda_2}}R\Big)R^2.
	$$
	Therefore, by the concentration theorem in Lemma 3, we have
	$$
	\mathbb{P}\Big(\boldsymbol{Z}\geq \tilde{D}R^2+	N_0\sqrt{\tilde{C}}\frac{R\sqrt{t}}{\sqrt{n}}
	+\frac{2N_0\tilde{C}t}{3n}\Big)\leq \exp[-t],\quad \forall \, t>0,
	$$
	where $\tilde{D}:=2N_0\delta_1C_0^2\delta_0+ 2N_0\frac{\mathcal{Q}_n(\sqrt{\lambda_2})}{\lambda_2}+N_0C_0\sqrt{\delta_0/2}\frac{\mathcal{Q}_n(\sqrt{\lambda_2})}{\lambda_1\sqrt{\lambda_2}}R$.
	Note that, by the spectral assumption (Assumption D(i)), it is easy to check that 
	$\mathcal{Q}_n(\sqrt{\lambda_2})=O(\frac{\lambda_2^{(1-\tau)/2}}{\sqrt{n}})$ and thus $\frac{\mathcal{Q}_n(\sqrt{\lambda_2})}{\lambda_2}=O(\frac{1}{\sqrt{n\lambda_2^{(1+\tau)}}}))=O(1/\sqrt{\kappa_1}) $ following the assumption $n\lambda_2^{(1+\tau)}\geq \kappa_1$. Similarly, we also have
	$\frac{\mathcal{Q}_n(\sqrt{\lambda_2})}{\lambda_1\sqrt{\lambda_2}}R=O(\sqrt{\frac{\lambda_2^{(1-\tau)}}{\kappa_1}})$ based on $R^2\leq \lambda_1$ and $\sqrt{n}\lambda_1\geq 1$.
	We now take $t=n(\delta_1'')^2\lambda_2$ and assume that $\lambda_2\leq c_1R^2$ with some constant $c_1$. Taking $\delta_1$ and $\delta_1''$ small enough but $\kappa_1$ large enough, such that
	$$
	\tilde{D}+N_0\sqrt{\tilde{C}}\delta_1''+2\sqrt{c_1}\tilde{C}D(\delta_1'')^2+\frac{2}{3}c_1N_0\tilde{C}
	(\delta_1'')^2\leq \delta_0.
	$$
	So that
	$$
	\mathbb{P}\Big(\sup_{g\in \mathcal{G}(R)}\big|\|g\|_n^2-\|g\|^2\big|\geq \delta_0R^2\Big)\leq \exp[-n(\delta_1'')^2\lambda_2].
	$$
\end{proof}

{\bf Appendix C. Proof for $\mathcal{T}_2(\delta_0,R)$}
	
To verify that the event $\mathcal{T}_2(\delta_0,R)$ occurs with high probability, 
we make use of an upper bound of the Gaussian process stated as follows.
\begin{lem}\label{Guassianprocess}
	With the same conditions as Lemma \ref{boundw}, we have that
	$$
	\sup_{g\in \mathcal{G}(R)} \Big|\frac{1}{n}\sum_{i=1}^n \epsilon_ig(X_i,T_i)\Big|
	\leq \delta_0R^2,
	$$
	with probability at least $1-\exp[-n(\delta_1'')^2\lambda_2]$.
\end{lem}

The Gaussian concentration inequality from Theorem 7.1 of \cite{Ledoux2001} is a useful tool in our refined analysis, which provides tighter bounds than the general sub-Gaussian cases. In particular, the super-norm bounds of random variables
are not needed, as opposed to Rademacher concentration inequality presented in  Lemma  \ref{talagrand}.  
\begin{lem}\label{talagrand1}
	Let $\mathbb{G}=\{G_t\}_{t\in T}$ be a centered Gaussian process indexed by a countable set $T$ such that
	$\sup_{t\in T}G_t<\infty$ almost surely. Then
	$$
	\mathbb{P}\Big(\sup_{t\in T}G_t\geq \mathbb{E}[\sup_{t\in T}G_t]+\sqrt{r}\Big)\leq \exp(-\frac{r}{2\sigma^2}),
	$$
	where $\sigma^2=\sup_{t\in T}\mathbb{E}[G_t^2]<\infty$.
\end{lem}

{\bf Proof of Lemma \ref{Guassianprocess}.}
Note that for any $g\in \mathcal{G}(R)$,
$$
\Big|\frac{1}{n}\sum_{i=1}^n \epsilon_ig(X_i,T_i)\Big|\leq \Big|\frac{1}{n}\sum_{i=1}^n \epsilon_if(T_i)\Big|+
\sup_{j}\Big|\frac{1}{n}\sum_{i=1}^n \epsilon_iX_{ij}\Big| \|\bbeta\|_1.
$$
On one hand, we conclude from the conclusion in \eqref{ramadercom}  below that
$$
\mathbb{E}\Big(\sup_{g\in \mathcal{G}(R)}\Big|\frac{1}{n}\sum_{i=1}^{n}\epsilon_if(T_i)\Big|\Big)\leq \frac{\mathcal{Q}_n(\sqrt{\lambda_2})}{\sqrt{\lambda_2}}R.
$$
%where $\tilde{\delta}:=\big(1+\frac{\Lambda_{\max}}{\Lambda_{\min}}\big)\mu$.
In addition, since $\epsilon_i$'s are the standard Gaussian variables and $|X_{ij}|\leq C_0$ by Assumption B,  Bernstein inequality is applied to yield
$$
\mathbb{E}\Big\|\frac{1}{n}\sum_{i=1}^{n}X_i \epsilon_i\Big\|_\infty\leq  C_0\sqrt{\log(2p)/n}.
$$ 
Thus, using  similar arguments to \eqref{parradm} and \eqref{squradem} yields that
\begin{align*}
\mathbb{E}\sup_{g\in \mathcal{G}(R)} \Big|\frac{1}{n}\sum_{i=1}^n \epsilon_ig(X_i,T_i)\Big|&\leq \frac{\mathcal{Q}_n(\sqrt{\lambda_2})}{\sqrt{\lambda_2}}R+C_0\sqrt{\frac{\delta_0\log(2p)}{2n}}
\frac{R^2}{\lambda_1}\\
&\leq \sqrt{\frac{c_1}{\kappa_1}}R^2+C_0\delta_1\sqrt{\delta_0/2}R^2,
\end{align*}
which we follow the same augments in the last proof part of Lemma \ref{boundw}. Observe that $\Big|\frac{1}{n}\sum_{i=1}^n \epsilon_ig(X_i,T_i)\Big|$ is a centered Gaussian process by Assumption A, and also check that  
$\sigma^2\leq \frac{1}{n}R^2$ in Lemma \ref{talagrand1}. Then, by the Gaussian concentration inequality with $r=2(\delta_1'')^2\lambda_2R^2$, we have 
$$
\sup_{g\in \mathcal{G}(R)} \Big|\frac{1}{n}\sum_{i=1}^n \epsilon_ig(X_i,T_i)\Big|
\leq \sqrt{c_1}(1/\sqrt{\kappa_1}+\sqrt{2}\delta_1'')R^2+C_0\delta_1\sqrt{\delta_0/2}R^2.
$$
As long as $\delta_1,\delta_1''$ are small sufficiently and $\kappa_1$ is properly large, we can obtain the desired result. 
$\Box$

\end{document}